%% file: main_arxiv.tex
\pdfoutput=1

\documentclass[10pt,twocolumn,letterpaper]{article}

\usepackage{cvpr}      %

\input{preamble_clean}
\renewcommand{\TODO}[1]{}\renewcommand{\chen}[1]{}\renewcommand{\HT}[1]{}
\usepackage{lipsum}

\definecolor{cvprblue}{rgb}{0.21,0.49,0.74}
\usepackage[pagebackref,breaklinks,colorlinks,allcolors=cvprblue]{hyperref}

\def\confName{CVPR}
\def\confYear{2027}

\def\name{Lyris\xspace}

\title{\name: Layout-Guided Long Video Generation with Persistent 3D Memory}
\author{
Chen Wang$^{1,2*}$ \quad Jiahui Huang$^{1}$ \quad Riccardo de Lutio$^{1}$ \quad Alex Trevithick$^{1}$ \quad Bryan Chu$^{1}$ \\
Jiawei Ren$^{1}$ \quad Laura Leal-Taix\'e$^{1}$ \quad Shengyu Huang$^{1}$ \quad Lingjie Liu$^{2}$ \quad Haithem Turki$^{1}$ \\
\vspace{3pt}
$^1$NVIDIA \quad $^2$University of Pennsylvania \\
\vspace{3pt}
{\small\url{https://research.nvidia.com/labs/sil/projects/lyris/}}
}

\begin{document}
\renewcommand{\thefootnote}{\fnsymbol{footnote}}
\maketitle

\begin{strip}
    \centering
    \vspace{-5em}
    \centering
    \includegraphics[width=\linewidth]{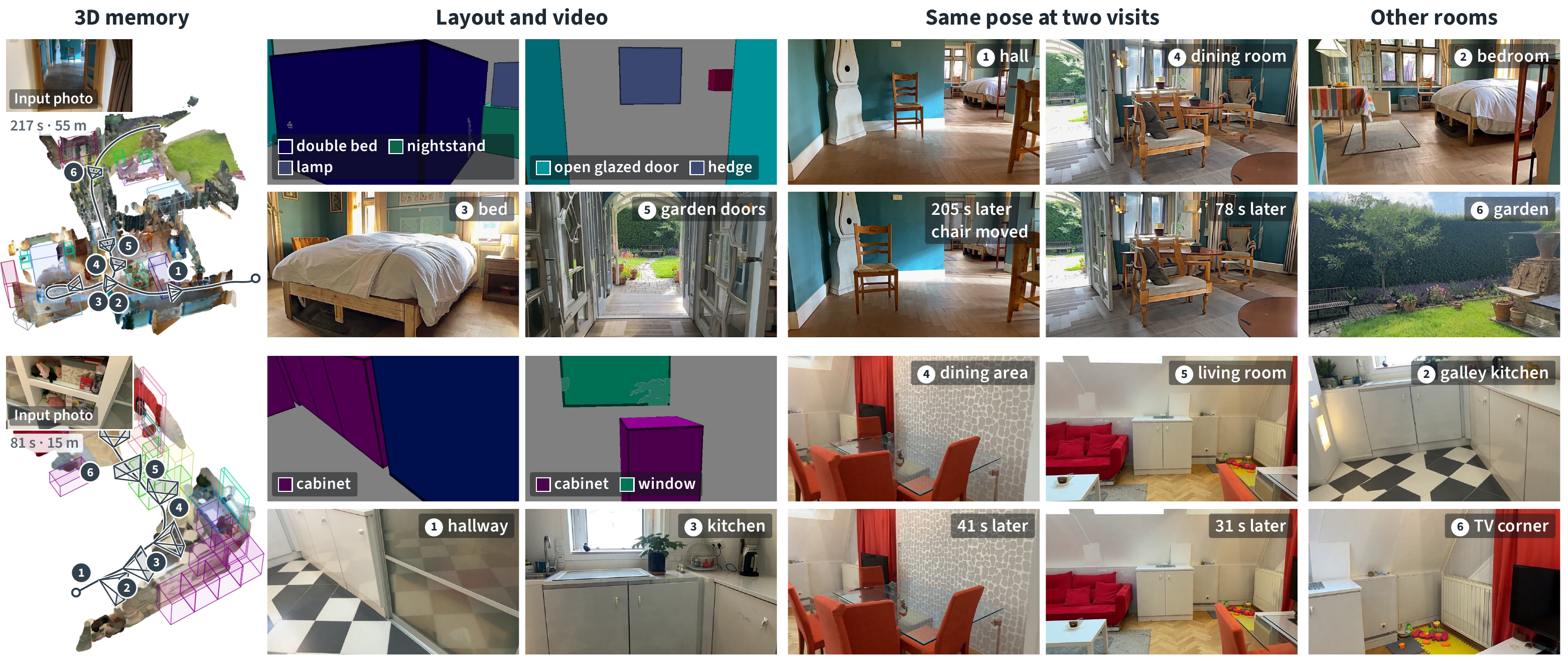}
    \vspace{-15pt}
    \captionof{figure}{
    From a single photo, \name generates long, camera-controlled walkthroughs that extend far beyond it. A vision-language model authors a 3D layout of labeled object bounding boxes for the video model to follow, and a user can edit it (as we do for the bedroom and garden doors, and to move the hall chair). A persistent point-cloud memory keeps rooms consistent upon return, even three minutes later.
    }
    \vspace{-10pt}
    \label{fig:teaser}
\end{strip}
\def\thefootnote{*}\footnotetext{Work done during an internship at NVIDIA.}
\def\thefootnote{\arabic{footnote}}

\input{sec_clean/0_abstract}    
\input{sec_clean/1_intro}
\begin{figure*}[t]
    \centering
    \includegraphics[width=\linewidth]{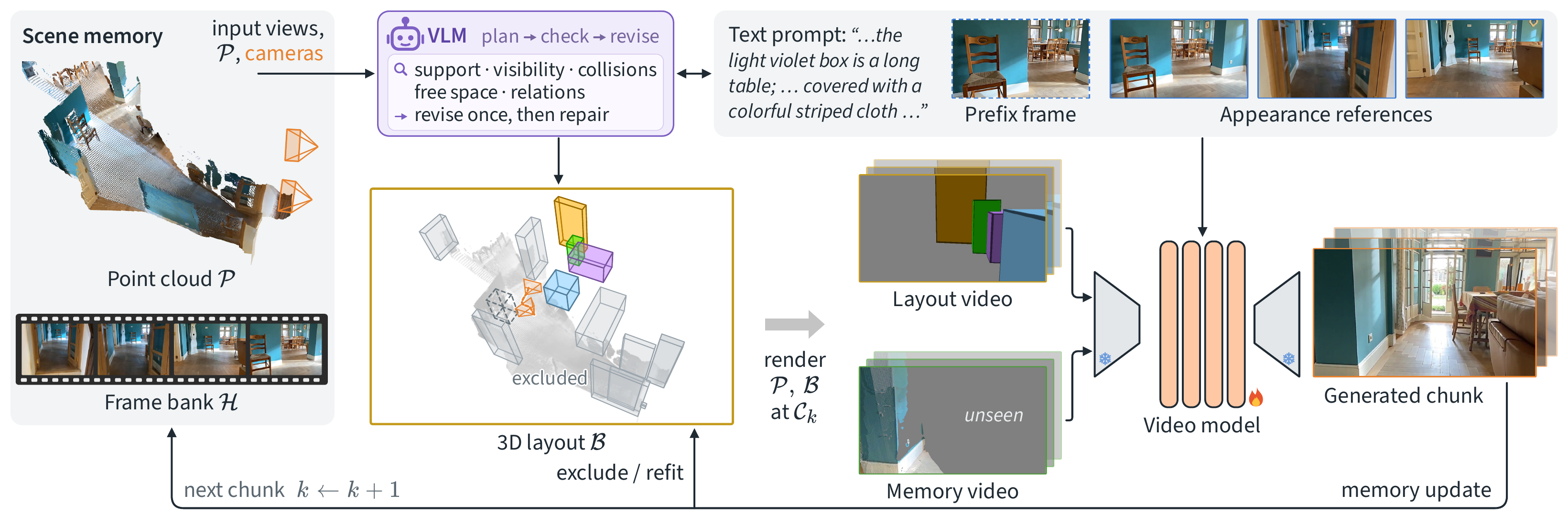}
    \vspace{-18pt}
    \caption{\textbf{Overview of \name.} To decide what lies beyond the input views, we use a vision-language model to plan a 3D layout $\mathcal{B}$ of labeled object boxes before each chunk $k$. We then render this layout and our point-cloud memory $\mathcal{P}$ at its cameras $\mathcal{C}_k$ to guide the video model. The memory video preserves what has been seen, while the layout video specifies what appears where beyond it and anchors the camera to its path. To remain consistent upon return, we reconstruct each generated chunk and write it into the memory.}
    \vspace{-11pt}
    \label{fig:pipeline}
\end{figure*}
\input{sec_clean/2_related-work}

\input{sec_clean/method_new}    %

\input{sec_clean/4_experiments}

\input{sec_clean/5_conclusion}
{
    \small
    \bibliographystyle{ieeenat_fullname}
    \bibliography{main}
}

\input{sec_clean/X_suppl}

\end{document}

%% file: preamble_clean.tex
\usepackage{cuted}
\usepackage{colortbl}
\definecolor{tabfirst}{rgb}{1, 0.7, 0.7}
\definecolor{tabsecond}{rgb}{1, 0.85, 0.7}
\definecolor{tabthird}{rgb}{1, 1, 0.7}

\newcommand{\TODO}[1]{\textbf{\color{red}[TODO: #1]}}
\newcommand{\chen}[1]{\textcolor{purple}{[Chen: #1]}}
\newcommand{\HT}[1]{\textcolor{orange!85!black}{[HT: #1]}}

\newcommand{\topic}[1]
{
\vspace{1pt}\noindent\textbf{#1}\ \ignorespaces
}

\newlength\paramargin
\newlength\figmargin

\newlength\secmargin
\newlength\figcapmargin
\newlength\tabcapmargin

\usepackage{multirow}
\usepackage{float}

\usepackage{tablefootnote}
\usepackage{pifont} %

\AtBeginDocument{\addtocontents{toc}{\protect\setcounter{tocdepth}{-1}}}

%% file: sec_clean/0_abstract.tex
\begin{abstract}
Exploring a scene from as little as a single photo requires generating a long, camera-controlled walkthrough that extends far beyond what was captured, yet remains consistent upon return.
Memory-based video generation methods preserve what they have already seen, but have no anchor beyond it, where the camera drifts from its path and a text prompt can name objects but not place them.
Layout-conditioned methods provide such an anchor and precise control over what appears where, but rely on layouts that are typically authored by hand.
Both limitations arise because, beyond what was captured, nothing anchors the video model unless a human authors one.
To overcome both, we propose \name, a long video generation framework that never leaves the video model without an anchor: a persistent point-cloud memory where the scene has been observed or generated, and a 3D layout that a vision-language model authors where it has not.
Our key insight is that the interface between the two models can be surprisingly coarse.
A layout of one labeled 3D box per object is simple enough for a vision-language model to author (and for a user to edit), yet expressive enough for a video model to follow once rendered along the camera path.
We demonstrate walkthroughs of over three minutes and 50 meters from a single photo, passing through doorways into rooms it never shows.
When evaluated on minute-long round trips across four datasets, \name outperforms state-of-the-art methods by a wide margin, reducing camera error by 3.8$\times$ and improving consistency upon return by 22\% on average.

\end{abstract}

%% file: sec_clean/1_intro.tex
\vspace{-10pt}
\section{Introduction}
\label{sec:intro}

Virtual touring~\cite{turki2022mega}, content creation, and simulation for physical AI~\cite{turki2026simuli, omnidreams2026} benefit from 3D environments that can be explored freely. Creating such environments from as little as a single photo requires generating a long, camera-controlled walkthrough beyond the captured view, passing through doorways and into unseen rooms. These regions must form a plausible extension of the observed scene and retain their layout and appearance when the camera returns.

Recent video generative models~\cite{wan2025wan, minimax2026minimaxh3, agarwal2026cosmos} can generate photorealistic clips from an image and a text prompt, but only seconds at a time.
Memory-based methods extend them to long rollouts by conditioning each new chunk on the history, through compressed or retrieved past views~\cite{yu2025context, li2025vmem, shen2026lyra, hong2025relic}, learned implicit representations~\cite{yu2026worldcrafter, wei2026geometry}, or explicitly reconstructed geometry~\cite{zhao2026spatia, chen2026anyrecon, lv2026octworld}.
This memory preserves what has already been seen but offers nothing beyond it.
There, the camera path reaches the model only as a pose encoding~\cite{shen2026lyra, yu2026mosaicmem}, a far weaker signal than rendered geometry~\cite{ren2025gen3c, cao2025uni3c}, and the camera drifts from its path as unseen regions accumulate.
The content of the unseen regions is left to the text prompt, which can name objects and relate them coarsely, but cannot specify where they belong or how one space connects to the next.
Layout-conditioned methods instead condition generation on an explicit scene layout, such as untextured geometry, object boxes, or maps~\cite{kim2025videofrom3d, bhat2026loosecontrolvideo, lu2024infinicube}, and offer precise control over the spatial arrangement of the objects.
However, these methods take the layout as an input, typically authored by hand, which does not scale to open-ended exploration.

To overcome both limitations, we propose \name, a long video generation framework in which a vision-language model authors the layout of unseen regions before a video model generates them.
Our key insight is that the interface between the two models can be surprisingly coarse: a 3D layout with one labeled bounding box per object~\cite{feng2023layoutgpt, yang2023holodeck, sun2024layoutvlm} is simple enough for a vision-language model to author (and for a user to edit), yet expressive enough for a video model to follow.
We fine-tune a pretrained video diffusion model to follow renderings of this layout, which specify what appears where and anchor the camera to its path, even in unseen regions.
To remain consistent upon return, we reconstruct each generated chunk and write it into a persistent point-cloud memory, which conditions every subsequent chunk alongside the layout.
The resulting loop between explicit 3D geometry and photorealistic video generation extends the walkthrough far beyond the photo, as the layout and the memory condition each chunk, and each generated chunk in turn extends the memory from which the next layout is planned.

We make the following contributions:
\begin{itemize}
    \item We present \name, a long video generation framework that pairs a 3D layout of unseen regions planned by a vision-language model with a persistent 3D memory of existing ones, enabling consistent 3D world exploration from as little as a single image without human authoring.
    \item We demonstrate that the layout is crucial for camera control in unseen regions, and that rendering it into the target views is far more effective than stating it in the text prompt.
    \item We outperform state-of-the-art methods on minute-long round trips across four datasets, reducing camera error by 3.8$\times$ and improving consistency upon return by 22\% on average.
\end{itemize}

%% file: sec_clean/2_related-work.tex
\section{Related Work}
\label{sec:related-work}

\topic{Camera-controlled video generation.}
Video diffusion models~\cite{ho2022video} generate photorealistic clips~\cite{blattmann2023stable, yang2025cogvideox, wu2025hunyuanvideo, wan2025wan} and increasingly accept multimodal conditions~\cite{seedance2026seedance, minimax2026minimaxh3, agarwal2026cosmos}.
To control the viewpoint, existing methods inject per-frame camera poses~\cite{wang2024motionctrl} or dense Pl\"ucker ray embeddings~\cite{he2024cameractrl, xu2024camco}, often through dedicated conditioning branches~\cite{bahmani2025vd3d, bahmani2025ac3d}, and extend beyond single clips by conditioning each new clip on frames of the preceding one~\cite{he2025cameractrl2}.
Geometry-guided approaches instead condition on point-cloud renderings at the target views~\cite{yu2024viewcrafter, ren2025gen3c, cao2025uni3c, yu2025trajectorycrafter}, couple implicit reconstruction of the input views with generation~\cite{lang2026roge}, or repair renderings of imperfect reconstructions~\cite{wu2025difix3d, de2026artifixer}.
These methods guide generation where geometry has been observed, but leave the content of unobserved regions entirely to the generative prior.
SpatialCrafter~\cite{fang2026spatialcrafter} instead generates the geometry of these regions with a 3D generator conditioned on the input image alone, whereas \name lays them out as labeled boxes that a vision-language model can reason and write and a user can edit.

\smallskip
\topic{Scene memory for long video generation.}
To generate long videos, memory-based methods condition each new frame or chunk on previously observed or generated content.
Many approaches compress this history into a limited token budget~\cite{zhang2026frame, hong2025relic, chandratreya2026millivid, wei2026geometry} or retrieve its most relevant parts~\cite{yu2025context, xiao2026worldmem, li2025vmem, yang2026decmem}.
However, the model must still infer how this history maps to the target view, which often fails under large viewpoint changes~\cite{shen2026lyra}.
A second line of work addresses this by aligning the history with the target camera before generation~\cite{chou2025captainsafari, shen2026lyra, yu2026mosaicmem, yu2026worldcrafter}, or by maintaining scene geometry that is rendered into the target views~\cite{ren2025gen3c, wu2026video, zhao2026spatia, zhang2026worldstereo, huang2025voyager, chen2026anyrecon, lv2026octworld, hyworld2026hyworld}.
Although rendering aligns the memory with each target view, errors can accumulate in the geometry itself~\cite{zhang2026worldstereo, lv2026octworld, hyworld2026hyworld}.
Other methods instead generate and simulate an editable latent 3D scene around the agent, albeit trained with ground-truth 3D supervision in a voxel game~\cite{garcin2026persist}, or sample all views jointly in a geometry-native latent space~\cite{lu2026gae}.
Whatever its form, however, a memory of past frames preserves what has already been seen but offers nothing beyond it.
There, the camera can only be conveyed through pose encodings~\cite{shen2026lyra, yu2026mosaicmem}, which control it far less precisely than rendered geometry~\cite{ren2025gen3c, cao2025uni3c}.
The content, in turn, is left to the text prompt.
\name also maintains an explicit point-cloud memory, but pairs it with a 3D layout of the objects that unseen regions should contain.
Rendered into the same target views, this layout anchors the camera where the memory is empty and leaves the appearance of these objects, but no longer their placement, to the generative prior.
As the generated views follow their requested cameras more closely, they also align more reliably with the memory, limiting the errors that accumulate in its geometry.

\smallskip
\topic{Layout-conditioned generation.}
Scene structure is an effective control signal when it is available in advance.
Generative renderers turn animated proxy meshes~\cite{cai2023generativerendering}, 3D tracking videos~\cite{gu2025das}, coarse geometry~\cite{kim2025videofrom3d}, or depth rendered by a game engine~\cite{ma2026closingtheloop} into realistic or stylized video.
Although some remain consistent even when a region is revisited~\cite{gu2025das, ma2026closingtheloop}, they all require this structure to be modeled or captured beforehand.
Looser controls ease this requirement by specifying mainly where content should appear, as boxes for images~\cite{bhat2023loosecontrol} or short video clips~\cite{wang2024boximator, wang2025cinemaster, bhat2026loosecontrolvideo}.
Layout-guided methods similarly generate entire indoor scenes from 3D layouts~\cite{fang2023ctrlroom, schult2023controlroom3d, yang2024scenecraft} and street scenes from road maps and object boxes~\cite{gao2023magicdrive, lu2024infinicube, omnidreams2026, agarwal2026cosmos} or street and height maps~\cite{deng2024streetscapes}.
However, these controls are typically authored by hand or taken from an existing scene or map, which does not scale to open-ended exploration.
\name instead conditions on a coarse object layout that a vision-language model writes for unseen regions, along with a persistent memory that keeps revisited regions consistent.

\smallskip
\topic{Language-driven scene planning.}
Large language and vision-language models have emerged as effective planners for visual content.
Many systems use them to lay out 3D scenes~\cite{feng2023layoutgpt, yang2023holodeck, bian2025holodeck2, ling2025scenethesis, sun2024layoutvlm, zhang2026scenepilot, duan2025latticeworld}, but these scenes inherit the synthetic appearance of the assets and engines that realize them.
Other methods instead direct generative models through language, describing the content to generate next~\cite{yu2023wonderjourney, erkoc2026worldagents} or iteratively refining prompts or plans for video generation~\cite{long2025vista, yang2026spiral}.
Although this avoids synthetic appearance, their plans remain natural language, which can name content but only coarsely describe where it belongs.
\name bridges the two: a vision-language model authors a 3D layout of labeled boxes, and a video diffusion model turns it into a photorealistic video.
Closest to our work, recent methods also guide video models with 3D boxes that a language model plans or programs~\cite{kizil2026lamp, wang2026worlddirector, huang2026programmable}.
However, these boxes direct individual objects and leave the scene around them to the video model, whereas ours lay out the scene beyond the capture.
Concurrent to our work, AgiWorld~\cite{fang2026agiworld} also plans unseen regions with a language-model agent, but conveys its layout through the text prompt, a strategy that we find far less effective than rendering the layout (\cref{sec:diagnostics}).
CoDeR~\cite{fang2026coder} instead re-creates the entire scene in code, which only approximates what the input image shows, whereas we keep the captured scene and lay out what lies beyond it.

%% file: sec_clean/method_new.tex
\vspace{-10pt}
\section{Method}
\label{sec:method}
Given one or a few input views, a camera path, and an optional text description of what each part of the path should reveal, our goal is to generate a long video that follows the path far beyond the observed regions, while remaining consistent when the camera returns.
Our solution is a chunk-wise autoregressive video model guided by a 3D layout of labeled object boxes, which a vision-language model plans for unseen regions, and by a persistent point-cloud memory and appearance references.
We describe our object layouts in \cref{sec:layout}, our scene memory in \cref{sec:memory}, and our video model and its training in \cref{sec:training}.
\Cref{fig:pipeline} illustrates our overall approach.

\subsection{Object Layouts}
\label{sec:layout}
\begin{figure*}[t]
    \centering
    \includegraphics[width=\linewidth]{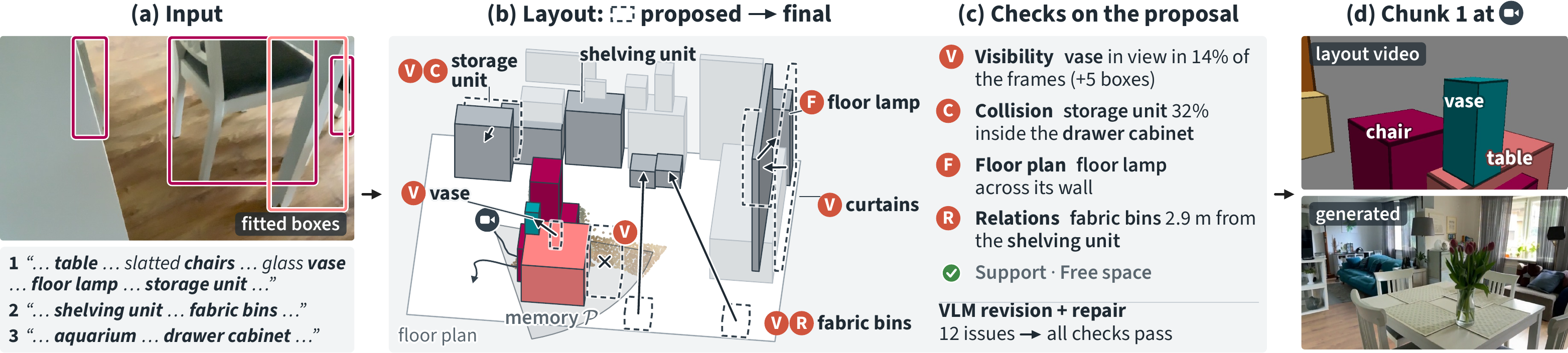}
    \vspace{-15pt}
    \caption{\textbf{Layout planning.} To keep the layout consistent with the observed scene, the vision-language model first fits boxes to the objects visible in the input view (\textbf{a}). It then adds a box for every object that the chunks should reveal, following their descriptions (\textbf{b}). To ensure that the layout is physically plausible and visible along the path, we verify the proposal with geometric checks (\textbf{c}) and revise the boxes they flag from their proposed positions (dashed in \textbf{b}) to their final ones. The video model then generates each object where its box places it, such as the vase, which was moved into view (\textbf{d}).
    }
    \label{fig:layout_planning}
    \vspace{-18pt}
\end{figure*}

\topic{Layout representation.}
We represent the layout as a set $\mathcal{B} = \{\mathbf{b}_i\}$ of labeled 3D bounding boxes, one per object, where each box $\mathbf{b}_i$ is parameterized by its center $\mathbf{c}_i \in \mathbb{R}^3$, size $\mathbf{s}_i \in \mathbb{R}^3$, rotation $\theta_i$ about the gravity axis, and a text label.
The boxes determine what unseen regions contain and where, and leave the shape and appearance of each object to the video model.
As each box is tied to a single object, a user can also change what is generated by moving or relabeling boxes before generation (\cref{fig:why_layouts}).

\topic{Planning.}
To let the layout build on what has already been generated, we use a vision-language model to plan the layout~\cite{feng2023layoutgpt, yang2023holodeck, sun2024layoutvlm} of what each chunk will reveal right before it is generated (\cref{fig:layout_planning}).
The planner receives the chunk's start frame, the point-cloud memory, the boxes of earlier chunks, and the chunk's cameras, and thus does not require the full camera path in advance.
To keep the layout consistent with the existing scene, it first estimates a rectangular floor plan that encloses the memory's points and the chunk's cameras and fits boxes to the objects visible in the start frame (\cref{fig:layout_planning}a).
It then determines which objects the chunk should reveal, following its description when one is given, and adds a box for each, placing it relative to the existing boxes and within the floor plan.

\topic{Verification.}
To ensure that the layout is physically plausible and visible along the path, we verify the proposal with geometric checks (\cref{fig:layout_planning}c).
The checks test whether each box is supported, lies within the floor plan and outside observed free space, does not collide with other boxes, is visible from the planned cameras, and satisfies the spatial relations of its description.
We then let the model revise its proposal once, given the violations and renderings of the layout from above and from the planned cameras, and resolve the remaining violations with a deterministic repair (\cref{fig:layout_planning}b).
We provide details in \cref{sec:supp_planning}.

\topic{Layout video and text prompt.}
We condition the video model on the layout through a layout video, which renders the boxes of objects yet to be generated from each chunk's cameras $\mathcal{C}_k$.
The layout video shows where each object belongs in every frame and anchors the camera to its path, even in unseen regions.
To indicate what each box contains, we render it in a color unique to its label and name the object of each color in the text prompt (\eg, ``the light violet box is a long table'' in \cref{fig:pipeline}).
\vspace{-3pt}
\subsection{Scene Memory}
\label{sec:memory}
As the video model keeps no state across chunks, we maintain a scene memory of everything observed or generated so far, as a colored point cloud $\mathcal{P}$ and a bank $\mathcal{H}$ of posed frames.
We initialize it by reconstructing the input views~\cite{kong2026moge, wang2026pi, pi3x}.

\topic{Memory video.}
To show the video model what the scene already contains, we render the point cloud $\mathcal{P}$ from each chunk's cameras $\mathcal{C}_k$, as in prior geometry-conditioned video models~\cite{ren2025gen3c, chen2026anyrecon, zhao2026spatia}, and provide the resulting memory video as a reference video.

\topic{Appearance references.}
As point renderings lack fine detail, we also provide a few frames of $\mathcal{H}$ as reference images.
We select the frames whose views overlap the most with the chunk's cameras, similar to Lyra2~\cite{shen2026lyra}, and provide details in \cref{sec:supp_retrieval}.

\topic{Memory update.}
To let later chunks build on what has been generated, we add each chunk to the memory once it is complete.
We reconstruct a subset of its frames with Pi3X~\cite{wang2026pi, pi3x} together with frames of the previous chunk, and align the reconstruction to the memory with a similarity transform fitted on these shared frames~\cite{umeyama1991least}.
To keep misaligned geometry out of the memory, we accept the update only if the aligned cameras deviate from the requested ones by at most $15^\circ$ in rotation and a quarter of their spread in translation.
We then add the new points to $\mathcal{P}$ and the frames to $\mathcal{H}$.
A chunk that fails this test remains in the video but leaves the memory unchanged.

\topic{Layout update.}
To keep the layout consistent with what has been generated, we then update it against the memory.
We exclude the boxes of objects that the point cloud now shows in full from the layout video, and refit the boxes of partially generated objects to their points, so that the remainder of each object continues the part that exists.
All other boxes remain as planned.

\subsection{Video Model}
\label{sec:training}
\label{sec:conditioning}
\topic{Model.}
We start from a pretrained bidirectional video diffusion model (MiniMax-H3 Ref2VA~\cite{minimax2026minimaxh3}), freeze its VAE and Qwen3-VL encoder~\cite{bai2025qwen3vl}, and fine-tune its transformer with LoRA~\cite{hu2022lora} via the original flow matching objective~\cite{lipman2023flow}.
As the model natively supports reference images and videos, we provide the layout and memory videos as reference videos and the appearance references as reference images. 

\topic{Prefix frame.}
To maintain temporal coherence across chunks, each chunk starts from a prefix frame: an input view for the first chunk, and the last frame of the previous chunk for every other.
The prefix frame also serves as an additional appearance reference.
\Cref{sec:supp_transitions} describes how we stabilize the transitions between chunks at inference.

\begin{figure*}[t]
    \centering
    \includegraphics[width=1.0\linewidth]{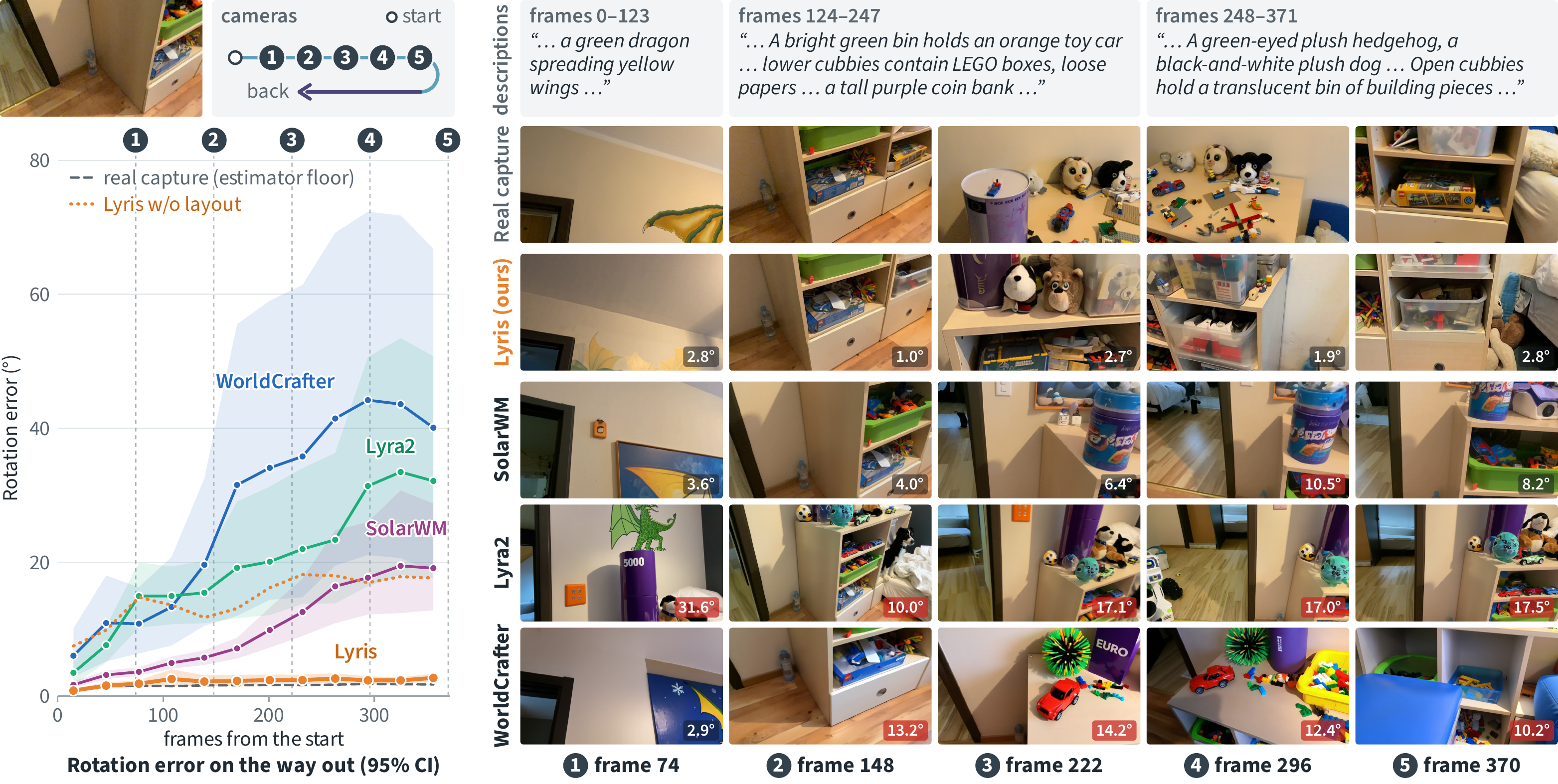}
    \vspace{-20pt}
    \caption{\textbf{Directed round trips.} Given the same start frame, cameras, and descriptions (\textbf{top}), we plot each method's rotation error along the way out over the 20 CA-1M scenes (\textbf{left}) and show five cameras of one round trip with each frame's error (\textbf{right}). The baselines drift once the path leaves the start view (by roughly 20--40$^\circ$ at its end), as does our model without its layout (dotted). \name is the only method to follow the path throughout, staying within $4^\circ$ of the requested cameras.
    }
    \label{fig:qualitative_comparison}
    \vspace{-8pt}
\end{figure*}

\topic{Camera embedding.}
The memory and layout videos convey the camera motion only where they show content.
To provide camera control even when both are entirely empty, we encode the cameras $\mathcal{C}_k$ as Pl\"ucker rays~\cite{he2024cameractrl} and add them to the tokens of the target and reference frames via a linear layer in each transformer block.
We zero-initialize these layers to ensure compatibility and stability with the pretrained initialization.

\topic{Training.}
We construct training examples from captured clips of ScanNet++~\cite{yeshwanth2023scannetplusplus} and DL3DV~\cite{ling2024dl3dv}, treating each clip as one chunk and its first frame as the prefix frame.
We render the memory video from geometry estimated from the input views instead of ground truth, so that the model must learn to handle the incomplete coverage and reconstruction artifacts that it encounters at inference.
We derive the layout video from the 3D instance annotations of ScanNet++, using the boxes of objects visible beyond the first frame, and omit it for DL3DV clips, which provide no such annotations.
We also drop the memory and layout videos independently with probability $0.2$ each (and the text prompt with probability $0.1$), so that a single model can generate from any subset of these conditions.

%% file: sec_clean/4_experiments.tex
\section{Experiments}
\label{sec:experiments}
As our goal is to extend captured scenes well beyond what was observed while remaining consistent upon return, we evaluate \name on long round-trip trajectories.
We first give every method the same detailed description of what lies ahead, and assess how well each generates it (\cref{sec:directed}).
Exploring a new scene usually comes without such descriptions, so in \cref{sec:exploration}, we only provide the caption of the start frame, and we assess whether planning what lies ahead with a vision-language model is
more effective than leaving it purely to the video model.
We validate the contribution of individual components and measure how closely \name follows its layouts in \cref{sec:diagnostics}.

\subsection{Setup}
\begin{table*}[t]
  \centering
  \caption{\textbf{Directed layouts.} All methods generate a 744-frame round trip from the same start frame, cameras, and descriptions. \name outperforms all baselines on every metric on both datasets, and follows the requested cameras more than four times as accurately as the next-best method (SolarWM-H3). We highlight the \protect\colorbox{tabfirst}{best}, \protect\colorbox{tabsecond}{second-best}, and \protect\colorbox{tabthird}{third-best} results.}
  \vspace{-9pt}
  \label{tab:directed}
  \setlength{\tabcolsep}{3.5pt}
  \resizebox{\linewidth}{!}{
  \begin{tabular}{@{}l|cccccccc|cccccccc@{}}
    \toprule
    & \multicolumn{8}{c|}{ScanNet++~\cite{yeshwanth2023scannetplusplus}} & \multicolumn{8}{c}{CA-1M~\cite{lazarow2025cubify}} \\
    Method & RotErr$\downarrow$ & TransErr$\downarrow$ & PSNR$\uparrow$ & SSIM$\uparrow$ & LPIPS$\downarrow$ & Revisit$\uparrow$ & Quality$\uparrow$ & Objects$\uparrow$ & RotErr$\downarrow$ & TransErr$\downarrow$ & PSNR$\uparrow$ & SSIM$\uparrow$ & LPIPS$\downarrow$ & Revisit$\uparrow$ & Quality$\uparrow$ & Objects$\uparrow$ \\
    \midrule
    AnyRecon~\cite{chen2026anyrecon} & 31.2 & 0.37 & 15.17 & 0.648 & 0.391 & \cellcolor{tabthird}0.582 & 65.6 & 0.378 & 45.9 & 0.81 & \cellcolor{tabthird}15.60 & \cellcolor{tabthird}0.648 & \cellcolor{tabsecond}0.457 & \cellcolor{tabthird}0.521 & 62.8 & 0.366 \\
    GEN3C~\cite{ren2025gen3c} & 12.8 & \cellcolor{tabsecond}0.24 & \cellcolor{tabsecond}17.51 & \cellcolor{tabsecond}0.729 & \cellcolor{tabsecond}0.339 & 0.459 & 47.9 & 0.693 & \cellcolor{tabthird}19.3 & \cellcolor{tabthird}0.46 & \cellcolor{tabsecond}16.49 & \cellcolor{tabsecond}0.667 & \cellcolor{tabthird}0.470 & 0.385 & 44.7 & 0.682 \\
    Lyra2~\cite{shen2026lyra} & \cellcolor{tabthird}10.5 & \cellcolor{tabthird}0.25 & \cellcolor{tabthird}16.53 & \cellcolor{tabthird}0.685 & \cellcolor{tabthird}0.359 & 0.489 & \cellcolor{tabsecond}69.2 & \cellcolor{tabsecond}0.820 & 19.8 & \cellcolor{tabthird}0.46 & 14.97 & 0.632 & 0.488 & 0.408 & \cellcolor{tabsecond}65.6 & \cellcolor{tabthird}0.736 \\
    SolarWM-H3~\cite{huang2026SolarWM-H3} & \cellcolor{tabsecond}7.7 & \cellcolor{tabthird}0.25 & 14.47 & 0.626 & 0.442 & 0.441 & 65.3 & 0.741 & \cellcolor{tabsecond}10.1 & \cellcolor{tabsecond}0.42 & 13.89 & 0.599 & 0.510 & 0.475 & 63.7 & 0.645 \\
    Spatia~\cite{zhao2026spatia} & 48.5 & 0.61 & 12.38 & 0.534 & 0.531 & 0.407 & 61.3 & 0.579 & 55.1 & 1.04 & 13.66 & 0.593 & 0.542 & 0.444 & 56.0 & 0.617 \\
    WorldCrafter~\cite{yu2026worldcrafter} & 10.9 & 0.29 & 13.31 & 0.603 & 0.454 & \cellcolor{tabsecond}0.656 & \cellcolor{tabthird}68.4 & \cellcolor{tabthird}0.809 & 27.6 & 0.58 & 13.75 & 0.594 & 0.512 & \cellcolor{tabsecond}0.629 & \cellcolor{tabthird}64.0 & \cellcolor{tabsecond}0.792 \\
    \hline
    \textbf{\name} & \cellcolor{tabfirst}1.3 & \cellcolor{tabfirst}0.04 & \cellcolor{tabfirst}20.19 & \cellcolor{tabfirst}0.764 & \cellcolor{tabfirst}0.251 & \cellcolor{tabfirst}0.826 & \cellcolor{tabfirst}74.8 & \cellcolor{tabfirst}0.899 & \cellcolor{tabfirst}2.2 & \cellcolor{tabfirst}0.07 & \cellcolor{tabfirst}18.50 & \cellcolor{tabfirst}0.696 & \cellcolor{tabfirst}0.395 & \cellcolor{tabfirst}0.755 & \cellcolor{tabfirst}70.4 & \cellcolor{tabfirst}0.908 \\
    \bottomrule
  \end{tabular}
  }
  \vspace{-15pt}
\end{table*}
\label{sec:setup}
\topic{Implementation.}
We fine-tune MiniMax-H3 Ref2VA~\cite{minimax2026minimaxh3} with rank-32 LoRA on its attention and feed-forward layers, using 124-frame clips at $832\times480$ resolution from $9{,}926$ DL3DV-10K~\cite{ling2024dl3dv} and $747$ ScanNet++~v2~\cite{yeshwanth2023scannetplusplus} scenes (the latter of which provide object annotations).
We train on clips that start from one to five input views for $1{,}500$ iterations using AdamW with a learning rate of $10^{-4}$ on 128 H100 GPUs.
We then add clips that continue from a previous one for another $500$ iterations (so that the model learns to build on its memory and retrieved references), and distill the resulting model into a few-step student via DMD2~\cite{yin2024improved}.
We use MoGe-3~\cite{kong2026moge} to lift single frames, Pi3X~\cite{wang2026pi, pi3x} to reconstruct multiple frames, GPT-6 to caption the training clips, and Claude Opus 5.5 to plan layouts.
At test time, we retrieve three appearance references per chunk and sample with eight steps and no guidance.
We provide more details in \cref{sec:supp_method}.

\begin{table*}[t]
  \centering
  \caption{\textbf{Autonomous exploration.} All methods receive only the start frame, its caption, and the cameras. \name outperforms all baselines on every metric on all four datasets, including the in-the-wild videos of SpatialVID. Metrics and highlighting follow \cref{tab:directed}, without Objects, which requires descriptions. Gray: \name with its layout planned for the whole camera path before generation, instead of chunk by chunk.
  }
  \vspace{-10pt}
  \label{tab:exploration}
  \setlength{\tabcolsep}{3.5pt}
  \resizebox{\linewidth}{!}{
  \begin{tabular}{@{}l|ccccccc|ccccccc@{}}
    \toprule
    & \multicolumn{7}{c|}{ScanNet++~\cite{yeshwanth2023scannetplusplus}} & \multicolumn{7}{c}{CA-1M~\cite{lazarow2025cubify}} \\
    Method & RotErr$\downarrow$ & TransErr$\downarrow$ & PSNR$\uparrow$ & SSIM$\uparrow$ & LPIPS$\downarrow$ & Revisit$\uparrow$ & Quality$\uparrow$ & RotErr$\downarrow$ & TransErr$\downarrow$ & PSNR$\uparrow$ & SSIM$\uparrow$ & LPIPS$\downarrow$ & Revisit$\uparrow$ & Quality$\uparrow$ \\
    \midrule
    AnyRecon~\cite{chen2026anyrecon} & 31.2 & 0.37 & 15.17 & 0.648 & 0.391 & \cellcolor{tabthird}0.582 & \cellcolor{tabthird}65.6 & 45.9 & 0.81 & \cellcolor{tabthird}15.60 & \cellcolor{tabthird}0.648 & \cellcolor{tabsecond}0.457 & \cellcolor{tabthird}0.521 & \cellcolor{tabsecond}62.8 \\
    GEN3C~\cite{ren2025gen3c} & 15.3 & \cellcolor{tabsecond}0.25 & \cellcolor{tabsecond}17.14 & \cellcolor{tabsecond}0.721 & \cellcolor{tabsecond}0.346 & 0.472 & 47.4 & 27.9 & \cellcolor{tabthird}0.47 & \cellcolor{tabsecond}16.37 & \cellcolor{tabsecond}0.670 & \cellcolor{tabthird}0.467 & 0.404 & 43.1 \\
    Lyra2~\cite{shen2026lyra} & \cellcolor{tabthird}14.1 & \cellcolor{tabthird}0.29 & \cellcolor{tabthird}15.56 & \cellcolor{tabthird}0.671 & \cellcolor{tabthird}0.387 & 0.525 & \cellcolor{tabsecond}66.9 & 25.7 & 0.51 & 15.11 & 0.638 & 0.490 & 0.430 & 60.4 \\
    SolarWM-H3~\cite{huang2026SolarWM-H3} & \cellcolor{tabsecond}9.7 & \cellcolor{tabthird}0.29 & 14.44 & 0.630 & 0.445 & 0.463 & 63.5 & \cellcolor{tabsecond}9.0 & \cellcolor{tabsecond}0.45 & 14.15 & 0.621 & 0.505 & 0.497 & \cellcolor{tabthird}61.1 \\
    Spatia~\cite{zhao2026spatia} & 54.5 & 0.69 & 12.69 & 0.566 & 0.523 & 0.423 & 58.0 & 60.3 & 1.07 & 13.74 & 0.596 & 0.542 & 0.471 & 51.5 \\
    WorldCrafter~\cite{yu2026worldcrafter} & 16.3 & 0.33 & 13.47 & 0.615 & 0.452 & \cellcolor{tabsecond}0.627 & 61.7 & \cellcolor{tabthird}24.8 & 0.60 & 13.59 & 0.590 & 0.518 & \cellcolor{tabsecond}0.577 & 57.7 \\
    \hline
    \textbf{\name} & \cellcolor{tabfirst}1.5 & \cellcolor{tabfirst}0.05 & \cellcolor{tabfirst}19.98 & \cellcolor{tabfirst}0.758 & \cellcolor{tabfirst}0.255 & \cellcolor{tabfirst}0.817 & \cellcolor{tabfirst}75.1 & \cellcolor{tabfirst}4.7 & \cellcolor{tabfirst}0.12 & \cellcolor{tabfirst}18.33 & \cellcolor{tabfirst}0.697 & \cellcolor{tabfirst}0.396 & \cellcolor{tabfirst}0.755 & \cellcolor{tabfirst}71.0 \\
    \name w/ full path & \textcolor{gray}{1.4} & \textcolor{gray}{0.04} & \textcolor{gray}{20.04} & \textcolor{gray}{0.761} & \textcolor{gray}{0.252} & \textcolor{gray}{0.835} & \textcolor{gray}{75.1} & \textcolor{gray}{1.9} & \textcolor{gray}{0.07} & \textcolor{gray}{18.30} & \textcolor{gray}{0.693} & \textcolor{gray}{0.398} & \textcolor{gray}{0.770} & \textcolor{gray}{71.9} \\
    \midrule
    & \multicolumn{7}{c|}{DL3DV~\cite{ling2024dl3dv}} & \multicolumn{7}{c}{SpatialVID~\cite{wang2025spatialvid}} \\
    \midrule
    AnyRecon~\cite{chen2026anyrecon} & 15.0 & 1.04 & 13.31 & 0.433 & \cellcolor{tabsecond}0.489 & \cellcolor{tabthird}0.565 & 61.1 & 47.8 & 2.06 & 14.07 & 0.515 & 0.436 & \cellcolor{tabthird}0.582 & 62.7 \\
    GEN3C~\cite{ren2025gen3c} & \cellcolor{tabthird}4.2 & \cellcolor{tabthird}0.57 & \cellcolor{tabthird}13.47 & \cellcolor{tabsecond}0.448 & 0.498 & 0.514 & 49.2 & 15.8 & 1.38 & \cellcolor{tabthird}14.81 & \cellcolor{tabthird}0.562 & \cellcolor{tabthird}0.417 & 0.468 & 53.6 \\
    Lyra2~\cite{shen2026lyra} & 7.8 & 0.64 & \cellcolor{tabsecond}13.48 & \cellcolor{tabthird}0.440 & \cellcolor{tabthird}0.491 & 0.561 & \cellcolor{tabthird}62.6 & \cellcolor{tabthird}15.3 & \cellcolor{tabthird}0.88 & \cellcolor{tabsecond}15.20 & \cellcolor{tabsecond}0.571 & \cellcolor{tabsecond}0.398 & 0.495 & \cellcolor{tabsecond}65.9 \\
    SolarWM-H3~\cite{huang2026SolarWM-H3} & 6.2 & 1.24 & 11.86 & 0.384 & 0.560 & 0.485 & 56.8 & \cellcolor{tabsecond}9.7 & \cellcolor{tabsecond}0.78 & 14.11 & 0.522 & 0.453 & 0.492 & 59.1 \\
    Spatia~\cite{zhao2026spatia} & 44.8 & 3.21 & 11.53 & 0.359 & 0.589 & 0.423 & 55.9 & 57.3 & 2.89 & 12.63 & 0.477 & 0.496 & 0.431 & 56.2 \\
    WorldCrafter~\cite{yu2026worldcrafter} & \cellcolor{tabsecond}2.4 & \cellcolor{tabsecond}0.26 & 12.02 & 0.376 & 0.532 & \cellcolor{tabsecond}0.719 & \cellcolor{tabsecond}65.7 & 22.6 & 1.21 & 12.94 & 0.487 & 0.468 & \cellcolor{tabsecond}0.661 & \cellcolor{tabthird}63.6 \\
    \hline
    \textbf{\name} & \cellcolor{tabfirst}0.9 & \cellcolor{tabfirst}0.08 & \cellcolor{tabfirst}15.60 & \cellcolor{tabfirst}0.488 & \cellcolor{tabfirst}0.409 & \cellcolor{tabfirst}0.792 & \cellcolor{tabfirst}71.5 & \cellcolor{tabfirst}7.2 & \cellcolor{tabfirst}0.29 & \cellcolor{tabfirst}16.73 & \cellcolor{tabfirst}0.584 & \cellcolor{tabfirst}0.358 & \cellcolor{tabfirst}0.777 & \cellcolor{tabfirst}72.3 \\
    \name w/ full path & \textcolor{gray}{1.0} & \textcolor{gray}{0.08} & \textcolor{gray}{15.49} & \textcolor{gray}{0.481} & \textcolor{gray}{0.414} & \textcolor{gray}{0.794} & \textcolor{gray}{71.4} & \textcolor{gray}{2.6} & \textcolor{gray}{0.24} & \textcolor{gray}{16.76} & \textcolor{gray}{0.586} & \textcolor{gray}{0.358} & \textcolor{gray}{0.795} & \textcolor{gray}{73.0} \\
    \bottomrule
  \end{tabular}
  }
  \vspace{-8pt}
\end{table*}

\begin{figure*}[t]
    \centering
    \includegraphics[width=1.0\linewidth]{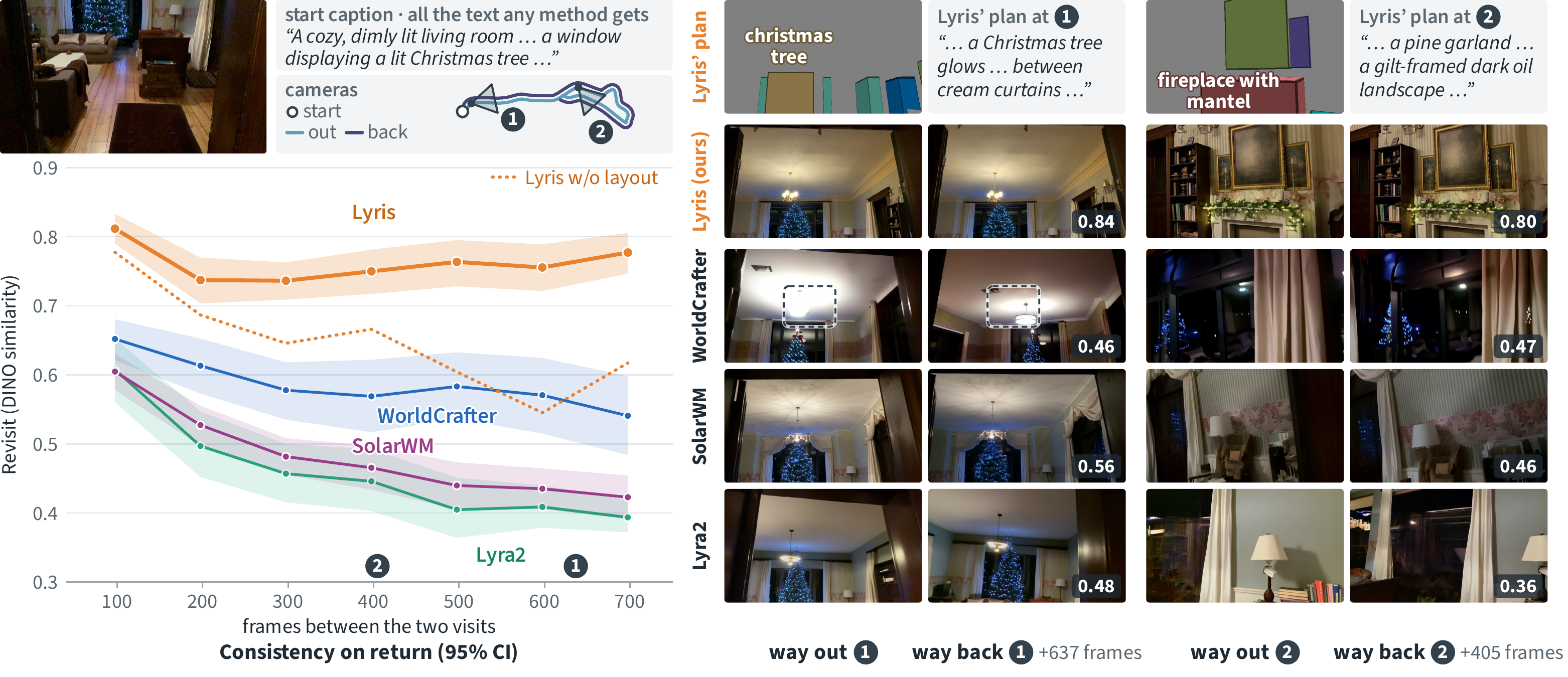}
    \vspace{-23pt}
    \caption{\textbf{Autonomous round trips.} Given only the start frame, its caption, and the cameras, we plot each method's consistency upon return (Revisit) against the frames between the two visits (\textbf{left}) and show two cameras on the way out and back (\textbf{right}). We average over the 40 CA-1M and ScanNet++ round trips and leave out views that the start frame shows, so that no method can copy its input. As the gap grows, the baselines return to shifted views or changed objects (such as WorldCrafter's ceiling light, dashed), and our model without its layout loses consistency as well (dotted). \name is the only method to return to the content it generated, even over 600 frames later.
    }
    \vspace{-17pt}
    \label{fig:long_rollout}
\end{figure*}

\topic{Round trips.}
Each round trip follows the cameras of a real capture for up to 372 frames on the way out and retraces them on the way back, for a total of up to 744 frames, or about one minute of video at 12\,fps.
On average, the start frame observes only 29\% of the pixels on the way out, leaving the rest to be generated, and the way back tests whether a method remembers what it generated there.
In the \emph{directed} setting (\cref{sec:directed}), every method receives a detailed description of each 124-frame window of the path, written by GPT-6 from the captured video, and in the \emph{autonomous} setting (\cref{sec:exploration}) only a caption of the start frame.
All evaluation scenes are held out from training.

\topic{Baselines.}
We compare \name to recent video generation methods that differ in how they leverage memory, namely AnyRecon~\cite{chen2026anyrecon}, GEN3C~\cite{ren2025gen3c}, and Spatia~\cite{zhao2026spatia}, which condition on point-cloud renderings, Lyra2~\cite{shen2026lyra}, which instead establishes dense correspondences to history frames through per-frame geometry, and WorldCrafter~\cite{yu2026worldcrafter}, which queries an implicit 3D-aware memory.
We also compare to SolarWM-H3~\cite{huang2026SolarWM-H3}, a camera-controlled world model that builds on MiniMax-H3 as we do.
All six baselines leave unseen regions directly to the video model.
We run the official code of each method with its default settings and the same start frame, cameras, and text as \name (except for AnyRecon, whose official code uses an empty prompt).

\topic{Metrics.}
We measure camera accuracy as the rotation and translation errors (RotErr, TransErr) between the input cameras and those that Depth Anything 3 (DA3)~\cite{lin2025depth} recovers from the generated video, aligned at the first frame and up to scale, and report the imaging quality score of VBench~\cite{huang2024vbench} (Quality).
As content generated in unseen regions need not match the ground truth, we compute PSNR, SSIM~\cite{wang2004image}, and LPIPS~\cite{zhang2018unreasonable} only over the pixels that the start frame observes, and measure consistency upon return as the DINOv2~\cite{oquab2024dinov2} patch similarity between the two frames generated at each camera, over all revisited cameras (Revisit).
As the descriptions of \cref{sec:directed} state what lies in unseen regions, we further report the fraction of the objects they name that a vision-language model finds in the generated video (Objects).

\subsection{Directed Layouts}
\label{sec:directed}
\topic{Datasets.} We evaluate on ScanNet++~\cite{yeshwanth2023scannetplusplus} and CA-1M~\cite{lazarow2025cubify} as real indoor captures with LiDAR depth and 3D object annotations, which we use in \cref{sec:diagnostics} to measure how closely \name follows its layouts.
We test on 20 held-out scenes of the former and the 20 longest landscape captures of the latter, which \name is not trained on.
We provide additional results on DL3DV-Evaluation~\cite{ling2024dl3dv} (not overlapping with DL3DV-10K scenes) and from sparse input views in \cref{sec:supp_dl3dv,sec:supp_sparse}.

\topic{Results.}
We summarize our results in \cref{tab:directed} and show a qualitative comparison in \cref{fig:qualitative_comparison}.
\name outperforms all baselines on every metric on both datasets, and follows the given cameras more than four times as accurately as the next-best method (SolarWM-H3).
The baselines drift from the path once it leaves the start view, as their memory offers nothing to anchor the camera in unseen regions, whereas \name renders its layout there (\cref{sec:diagnostics}).
Among the baselines, GEN3C best preserves the observed content through its explicit point cloud and WorldCrafter is the most consistent upon return through its implicit memory, but only our method achieves both by pairing a point cloud with retrieved appearance references.

\begin{figure*}[htbp]
    \centering
    \includegraphics[width=\linewidth]{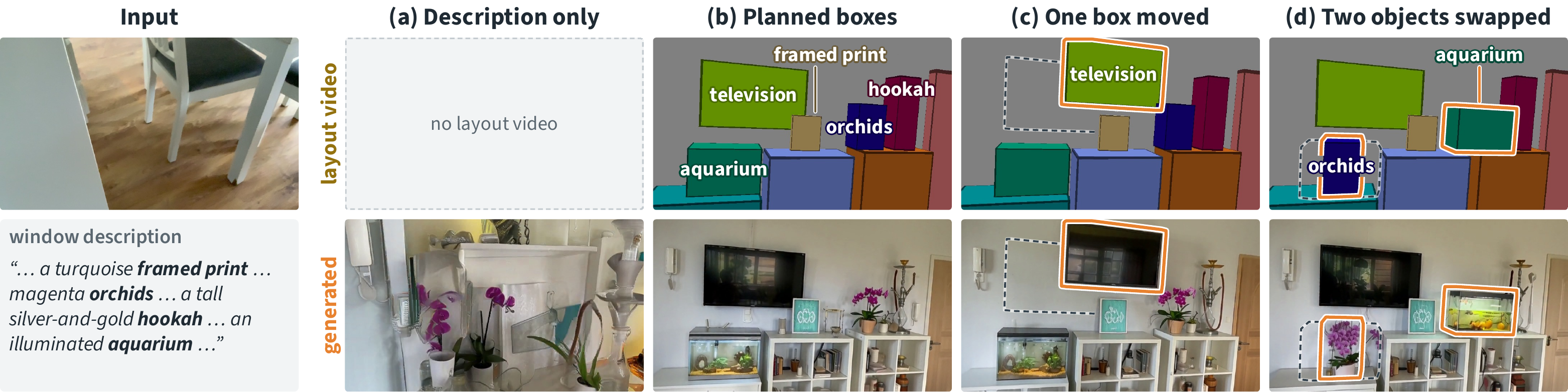}
    \vspace{-17pt}
    \caption{\textbf{Layout control.} We edit the plan of \cref{fig:layout_planning} and regenerate with the same inputs and seed. Without a layout, the model places the described objects itself and drifts from the requested camera (\textbf{a}), while with the plan each object appears in its box (\textbf{b}). Moving the television's box moves the television and leaves its former place (dashed) empty (\textbf{c}), and swapping the aquarium and the orchids in the layout and the descriptions swaps them (\textbf{d}), which lets a user edit the scene before it is generated.}
    \label{fig:why_layouts}
    \vspace{-18pt}
\end{figure*}

\subsection{Autonomous Exploration}
\label{sec:exploration}
\topic{Datasets.}
We evaluate on the ScanNet++ and CA-1M scenes of \cref{sec:directed} along with DL3DV-Evaluation~\cite{ling2024dl3dv} as a benchmark of diverse indoor and outdoor scenes, and SpatialVID~\cite{wang2025spatialvid} as in-the-wild videos that \name is not trained on.
We again test on 20 scenes of each and provide additional details in \cref{sec:supp_scenes}.

\topic{Results.}
We present quantitative results, using the same metrics as \cref{tab:directed}, in \cref{tab:exploration}, and qualitative results in \cref{fig:long_rollout}.
\name outperforms all baselines on every metric on all four datasets.
Even without the descriptions of \cref{sec:directed}, it outperforms every baseline that receives them on every metric (\cref{tab:directed}), as its planner decides what lies ahead.
\name also generalizes to the in-the-wild videos of SpatialVID, where it outperforms SolarWM-H3, which is trained on this dataset, on every metric.
When the full camera path is available before generation, planning the layout for the whole path at once (\name w/ full path) further improves camera accuracy on CA-1M and SpatialVID, following the latter's cameras more than three times as accurately as SolarWM-H3.
Averaged over the six blocks of \cref{tab:directed,tab:exploration}, \name reduces the rotation error of each block's best baseline by 3.8$\times$ and improves on its Revisit by 22\%.

\subsection{Ablation and Diagnostics}
\label{sec:diagnostics}
\topic{Components.}
We ablate our design decisions by individually omitting the major components of our method: the layout, the memory's point cloud, and the appearance references.
We summarize our results in \cref{tab:diagnostics}.
The layout is crucial, as omitting it increases the rotation error roughly three- to sevenfold and lowers consistency upon return, both with or without a description.
Stating the layout only in the prompt, as the planner's natural-language description of each chunk, instead of rendering it is far less effective, with a 2.8$\times$ higher rotation error.
Omitting the memory's point cloud degrades every metric, whereas omitting the appearance references mainly lowers consistency upon return.
\name also remains ahead of all baselines in camera accuracy and consistency upon return with a smaller planner (Claude Haiku 5.5), an open-weight one (Qwen3.8-Flash-Next), or the smaller Cosmos3-Nano~\cite{agarwal2026cosmos} as its video model.
We provide these results in \cref{sec:supp_planners,sec:supp_video_model}.

\begin{table}[t]
    \centering
    \caption{\textbf{Ablation study.} We evaluate on CA-1M and ScanNet++ round trips. The layout is crucial for camera accuracy with or without a description, and far more effective when rendered than when stated in the prompt.}
    \vspace{-10pt}
    \label{tab:diagnostics}
    \footnotesize
    \setlength{\tabcolsep}{0pt}
    \begin{tabular*}{\linewidth}{@{\extracolsep{\fill}}lccc@{}}
        \toprule
        & RotErr$\downarrow$ & PSNR$\uparrow$ & Revisit$\uparrow$ \\
        \midrule
        \multicolumn{4}{@{}l}{\emph{CA-1M, directed}} \\
        Full method & \textbf{2.2} & \textbf{18.50} & \textbf{0.755} \\
        \addlinespace[2pt]
        w/o layout & 14.6 & 18.00 & 0.653 \\
        w/o memory point cloud & 9.0 & 16.18 & 0.564 \\
        w/o appearance references & 2.9 & 18.08 & 0.707 \\
        \midrule
        \multicolumn{4}{@{}l}{\emph{CA-1M, autonomous}} \\
        Full method & \textbf{4.7} & \textbf{18.33} & \textbf{0.755} \\
        w/o layout & 12.2 & 17.99 & 0.670 \\
        Layout as text & 13.0 & 17.97 & 0.632 \\
        \midrule
        \multicolumn{4}{@{}l}{\emph{ScanNet++, directed}} \\
        Full method & \textbf{1.3} & \textbf{20.19} & \textbf{0.826} \\
        w/o layout & 5.3 & 20.12 & 0.785 \\
        \midrule
        \multicolumn{4}{@{}l}{\emph{ScanNet++, autonomous}} \\
        Full method & \textbf{1.5} & \textbf{19.98} & \textbf{0.817} \\
        w/o layout & 5.4 & 19.84 & 0.782 \\
        \bottomrule
    \end{tabular*}
    \vspace{-8pt} %
\end{table}

\begin{table}[t]
    \centering
    \caption{\textbf{Layout adherence.} We score the directed round trips of \cref{tab:directed}, generated with and without a layout, against the same boxes. \name places objects where its layout specifies, and an annotated layout thus raises the recall of the real objects.}
    \vspace{-10pt}
    \label{tab:layout}
    \footnotesize
    \setlength{\tabcolsep}{0pt}
    \begin{tabular*}{\linewidth}{@{\extracolsep{\fill}}lcc@{}}
        \toprule
        & w/o layout & w/ layout \\
        \midrule
        \multicolumn{3}{@{}l}{\emph{ScanNet++}} \\
        Adherence$\uparrow$ (planned layout) & 0.388 & 0.479 \\
        Adherence$\uparrow$ (annotated layout) & 0.456 & 0.648 \\
        Recall$\uparrow$ (annotated layout) & 0.341 & 0.437 \\
        \midrule
        \multicolumn{3}{@{}l}{\emph{CA-1M}} \\
        Adherence$\uparrow$ (planned layout) & 0.375 & 0.472 \\
        Adherence$\uparrow$ (annotated layout) & 0.477 & 0.583 \\
        Recall$\uparrow$ (annotated layout) & 0.324 & 0.511 \\
        \bottomrule
    \end{tabular*}
    \vspace{-15pt}
\end{table}

\topic{Layout control.}
To test whether the generation follows its layout, we edit a plan and regenerate with the same inputs and seed (\cref{fig:why_layouts}).
Moving a box moves its object and leaves its former place empty, and swapping two objects in the plan swaps them in the video, which lets a user edit the scene before it is generated.
Over the round trips of \cref{tab:directed}, we further measure the fraction of each box that the generated surface fills (adherence), for the objects that the start frame does not show (\cref{tab:layout}).
\name fills the boxes of its planned layout about a quarter more than when it generates without a layout.
Given the annotated layout of the real scene instead, \name follows it as well, and its content covers more of the real objects than without a layout (Recall).

\topic{Timing.}
Generating a 744-frame round trip takes about 28 minutes on a single H100 GPU, and planning its layout chunk by chunk adds a median of 10 minutes in the directed setting, for about 38 minutes end to end.

%% file: sec_clean/5_conclusion.tex
\vspace{-5pt}
\section{Conclusion}
\label{conclusion}

We presented \name, a long video generation framework that pairs a persistent point-cloud memory with a 3D layout that a vision-language model authors for unseen regions.
The resulting loop preserves the existing scene while extending it according to the layout, and provides an explicit geometric interface for specifying what is generated.
On minute-long round trips across four datasets, \name reduces the camera error of state-of-the-art methods by 3.8$\times$.

\noindent\textbf{Limitations.} Our method depends on the quality of the estimated geometry, as depth and alignment errors can accumulate in the memory and affect subsequent chunks.
Our model still requires multiple steps of diffusion sampling and is not yet interactive.

%% file: sec_clean/X_suppl.tex
\clearpage
\setcounter{page}{1}
\maketitlesupplementary
\ifdefined\putbib\begin{bibunit}[ieeenat_fullname]\fi  %
\appendix  %
\noindent We show videos of long walkthroughs, comparisons with the baselines, and layout edits on our project page: {\small\url{https://research.nvidia.com/labs/sil/projects/lyris/}}.

\addtocontents{toc}{\protect\setcounter{tocdepth}{2}}
{\renewcommand*\numberline[1]{#1\hskip 1em}\tableofcontents}  %

\section{Additional Results}
\label{sec:supp_results}

\subsection{Directed Round Trips on DL3DV}
\label{sec:supp_dl3dv}
\begin{table}[htbp]
  \centering
  \caption{\textbf{Directed round trips on DL3DV.} All methods generate a round trip of 662 to 730 frames on 20 DL3DV-Evaluation~\cite{ling2024dl3dv} scenes from the same start frame, cameras, and descriptions. \name outperforms all baselines on every metric. Metrics and highlighting follow \cref{tab:directed}, without Objects.}
  \label{tab:dl3dv_rt}
  \vspace{-10pt}
  \setlength{\tabcolsep}{3.5pt}
  \resizebox{\linewidth}{!}{
  \begin{tabular}{@{}l|ccccccc@{}}
    \toprule
    Method & RotErr$\downarrow$ & TransErr$\downarrow$ & PSNR$\uparrow$ & SSIM$\uparrow$ & LPIPS$\downarrow$ & Revisit$\uparrow$ & Quality$\uparrow$ \\
    \midrule
    AnyRecon~\cite{chen2026anyrecon} & 15.0 & 1.04 & 13.31 & 0.433 & \cellcolor{tabsecond}0.489 & 0.565 & 61.1 \\
    GEN3C~\cite{ren2025gen3c} & \cellcolor{tabthird}4.4 & 0.56 & \cellcolor{tabsecond}13.59 & \cellcolor{tabsecond}0.449 & 0.496 & 0.508 & 50.0 \\
    Lyra2~\cite{shen2026lyra} & 5.7 & \cellcolor{tabthird}0.53 & \cellcolor{tabthird}13.48 & \cellcolor{tabthird}0.436 & \cellcolor{tabthird}0.495 & \cellcolor{tabthird}0.569 & \cellcolor{tabthird}63.3 \\
    SolarWM-H3~\cite{huang2026SolarWM-H3} & 5.6 & 1.22 & 11.90 & 0.385 & 0.558 & 0.482 & 56.4 \\
    Spatia~\cite{zhao2026spatia} & 43.9 & 3.13 & 11.36 & 0.339 & 0.594 & 0.430 & 58.9 \\
    WorldCrafter~\cite{yu2026worldcrafter} & \cellcolor{tabsecond}2.0 & \cellcolor{tabsecond}0.24 & 11.98 & 0.370 & 0.532 & \cellcolor{tabsecond}0.718 & \cellcolor{tabsecond}66.9 \\
    \hline
    \textbf{\name} & \cellcolor{tabfirst}0.9 & \cellcolor{tabfirst}0.08 & \cellcolor{tabfirst}15.61 & \cellcolor{tabfirst}0.487 & \cellcolor{tabfirst}0.407 & \cellcolor{tabfirst}0.791 & \cellcolor{tabfirst}71.7 \\
    \bottomrule
  \end{tabular}
  }
  \vspace{-10pt}
\end{table}
We repeat the comparison of \cref{tab:directed} on the 20 indoor and outdoor DL3DV-Evaluation~\cite{ling2024dl3dv} scenes of \cref{tab:exploration} and summarize our results in \cref{tab:dl3dv_rt}.
\name again outperforms all baselines on every metric, and follows the given cameras more than twice as accurately as the next-best method (WorldCrafter).

\subsection{Smaller and Open-Weight Planners}
\label{sec:supp_planners}
\begin{table*}[t]
    \centering
    \caption{\textbf{Planners.} We replace the planner on the directed CA-1M round trips of \cref{tab:directed}, and report the metrics of that table along with the fractions of the described objects and spatial relations that its final layout places (Placed) and satisfies (Relations). Even with a smaller or an open-weight planner, \name outperforms the best baseline of each metric by a wide margin.}
    \vspace{-10pt}
    \label{tab:planners}
    \small
    \setlength{\tabcolsep}{0pt}
    \begin{tabular*}{\linewidth}{@{\extracolsep{\fill}}lcccccccccc@{}}
        \toprule
        Planner & RotErr$\downarrow$ & TransErr$\downarrow$ & PSNR$\uparrow$ & SSIM$\uparrow$ & LPIPS$\downarrow$ & Revisit$\uparrow$ & Quality$\uparrow$ & Objects$\uparrow$ & Placed$\uparrow$ & Relations$\uparrow$ \\
        \midrule
        Claude Opus 5.5 (\name) & 2.2 & 0.07 & 18.50 & 0.696 & 0.395 & 0.755 & 70.4 & 0.908 & 0.745 & 0.559 \\
        Claude Haiku 5.5 & 2.6 & 0.08 & 18.36 & 0.697 & 0.396 & 0.744 & 70.7 & 0.912 & 0.623 & 0.347 \\
        Qwen3.8-Flash-Next & 2.4 & 0.08 & 18.13 & 0.689 & 0.402 & 0.752 & 70.7 & 0.894 & 0.674 & 0.434 \\
        \midrule
        Best baseline & 10.1 & 0.42 & 16.49 & 0.667 & 0.457 & 0.629 & 65.6 & 0.792 & -- & -- \\
        \bottomrule
    \end{tabular*}
\end{table*}
We replace Claude Opus 5.5 with the smaller Claude Haiku 5.5 and with the open-weight Qwen3.8-Flash-Next\footnote{\texttt{Qwen/Qwen3.8-Flash-Next} on Hugging Face, revision \texttt{812d96e1}, served with vLLM at its released precision.} on the directed CA-1M round trips of \cref{tab:directed}, and summarize our results in \cref{tab:planners}.
With either planner, \name still outperforms all baselines on every metric of \cref{tab:directed}, and follows the cameras roughly four times as accurately as the best of them.
A stronger planner mainly produces a more complete layout, which places more of the described objects and satisfies more of their stated relations.

\subsection{Cosmos3-Nano as the Video Model}
\label{sec:supp_video_model}
\begin{table}[t]
  \centering
  \caption{\textbf{Video model.} We replace MiniMax-H3 with Cosmos3-Nano~\cite{agarwal2026cosmos} on the round trips of \cref{tab:directed,tab:exploration,tab:dl3dv_rt}, and list the best baseline of each metric. Metrics follow \cref{tab:exploration}. \name remains ahead of the baselines on every metric with this smaller model.}
  \vspace{-10pt}
  \label{tab:video_model}
  \setlength{\tabcolsep}{3.5pt}
  \resizebox{\linewidth}{!}{
  \begin{tabular}{@{}l|ccccccc@{}}
    \toprule
    Method & RotErr$\downarrow$ & TransErr$\downarrow$ & PSNR$\uparrow$ & SSIM$\uparrow$ & LPIPS$\downarrow$ & Revisit$\uparrow$ & Quality$\uparrow$ \\
    \midrule
    \multicolumn{8}{@{}l}{\emph{Directed, ScanNet++}} \\
    Best baseline & 7.7 & 0.24 & 17.51 & 0.729 & 0.339 & 0.656 & 69.2 \\
    \name with Cosmos3-Nano & \textbf{1.4} & \textbf{0.05} & \textbf{19.96} & \textbf{0.753} & \textbf{0.245} & \textbf{0.807} & \textbf{71.9} \\
    \midrule
    \multicolumn{8}{@{}l}{\emph{Directed, CA-1M}} \\
    Best baseline & 10.1 & 0.42 & 16.49 & 0.667 & 0.457 & 0.629 & 65.6 \\
    \name with Cosmos3-Nano & \textbf{4.2} & \textbf{0.17} & \textbf{17.95} & \textbf{0.687} & \textbf{0.398} & \textbf{0.740} & \textbf{65.7} \\
    \midrule
    \multicolumn{8}{@{}l}{\emph{Directed, DL3DV}} \\
    Best baseline & 2.0 & 0.24 & 13.59 & 0.449 & 0.489 & 0.718 & 66.9 \\
    \name with Cosmos3-Nano & \textbf{0.8} & \textbf{0.07} & \textbf{15.38} & \textbf{0.496} & \textbf{0.397} & \textbf{0.803} & \textbf{70.9} \\
    \midrule
    \multicolumn{8}{@{}l}{\emph{Autonomous, ScanNet++}} \\
    Best baseline & 9.7 & 0.25 & 17.14 & 0.721 & 0.346 & 0.627 & 66.9 \\
    \name with Cosmos3-Nano & \textbf{1.4} & \textbf{0.04} & \textbf{19.70} & \textbf{0.751} & \textbf{0.249} & \textbf{0.818} & \textbf{72.2} \\
    \midrule
    \multicolumn{8}{@{}l}{\emph{Autonomous, CA-1M}} \\
    Best baseline & 9.0 & 0.45 & 16.37 & 0.670 & 0.457 & 0.577 & 62.8 \\
    \name with Cosmos3-Nano & \textbf{4.4} & \textbf{0.09} & \textbf{18.02} & \textbf{0.696} & \textbf{0.392} & \textbf{0.745} & \textbf{67.3} \\
    \midrule
    \multicolumn{8}{@{}l}{\emph{Autonomous, DL3DV}} \\
    Best baseline & 2.4 & 0.26 & 13.48 & 0.448 & 0.489 & 0.719 & 65.7 \\
    \name with Cosmos3-Nano & \textbf{0.8} & \textbf{0.08} & \textbf{15.35} & \textbf{0.496} & \textbf{0.398} & \textbf{0.799} & \textbf{70.8} \\
    \midrule
    \multicolumn{8}{@{}l}{\emph{Autonomous, SpatialVID}} \\
    Best baseline & 9.7 & 0.78 & 15.20 & 0.571 & 0.398 & 0.661 & 65.9 \\
    \name with Cosmos3-Nano & \textbf{6.1} & \textbf{0.51} & \textbf{16.63} & \textbf{0.589} & \textbf{0.350} & \textbf{0.768} & \textbf{70.9} \\
    \bottomrule
  \end{tabular}
  }
\end{table}
To validate the effectiveness of our pipeline, we replace MiniMax-H3, whose transformer has 33B parameters, with the smaller Cosmos3-Nano~\cite{agarwal2026cosmos}, whose generator tower has 8B (\cref{tab:video_model}).
We fully fine-tune this tower in the same two stages and on as many clips as \name (\cref{sec:setup}), conditioning on the memory video, the layout video, and the appearance references, and distill it with DMD2.
On the round trips of \cref{tab:directed,tab:exploration,tab:dl3dv_rt}, it still outperforms all baselines on every metric of \cref{tab:exploration}.

\subsection{Sparse Input Views}
\label{sec:supp_sparse}
\begin{table}[t]
  \centering
  \caption{\textbf{Sparse input views.} We generate each path from one, three, or five input views, and compare it to the captured video over the full frame. The views are spread along the path (left, scored on all other frames) or lie within its first 40\% (right, scored on the remaining 60\%). Additional views improve fidelity on every dataset, including beyond the last view.}
  \vspace{-10pt}
  \label{tab:sparse}
  \setlength{\tabcolsep}{3.5pt}
  \resizebox{\linewidth}{!}{
  \begin{tabular}{@{}ll|ccc|ccc@{}}
    \toprule
    & & \multicolumn{3}{c|}{Views along the path} & \multicolumn{3}{c}{Views in the first 40\%} \\
    Dataset & Views & PSNR$\uparrow$ & SSIM$\uparrow$ & LPIPS$\downarrow$ & PSNR$\uparrow$ & SSIM$\uparrow$ & LPIPS$\downarrow$ \\
    \midrule
    ScanNet++ & 1 & 13.86 & 0.573 & 0.456 & 12.39 & 0.529 & 0.510 \\
                & 3 & 16.46 & 0.632 & 0.361 & 14.35 & 0.579 & 0.432 \\
                & 5 & \textbf{17.34} & \textbf{0.652} & \textbf{0.331} & \textbf{14.96} & \textbf{0.591} & \textbf{0.413} \\
    \midrule
    DL3DV & 1 & 14.84 & 0.487 & 0.426 & 14.31 & 0.472 & 0.441 \\
                & 3 & 16.74 & 0.545 & 0.350 & 16.29 & 0.531 & 0.359 \\
                & 5 & \textbf{17.38} & \textbf{0.564} & \textbf{0.325} & \textbf{16.51} & \textbf{0.536} & \textbf{0.352} \\
    \midrule
    SpatialVID & 1 & 13.80 & 0.533 & 0.464 & 12.62 & 0.487 & 0.513 \\
                & 3 & 16.14 & 0.604 & 0.360 & 14.77 & 0.554 & 0.421 \\
                & 5 & \textbf{17.64} & \textbf{0.641} & \textbf{0.304} & \textbf{14.83} & \textbf{0.558} & \textbf{0.416} \\
    \bottomrule
  \end{tabular}
  }
\end{table}
Although our main experiments start from a single photo, \name also supports multiple posed views as input.
We reconstruct these views into the initial point cloud of our scene memory and add them to its frame bank, so that each chunk can retrieve them as appearance references.
To measure the impact of additional views, we generate the way out of the ScanNet++, DL3DV, and SpatialVID round trips of \cref{tab:exploration} from one, three, or five views in the directed setting of \cref{sec:directed}.
We summarize our results in \cref{tab:sparse} and show examples on our project page.
Five views spread along the path improve PSNR over a single view by $2.5$ to $3.8$\,dB on every dataset (and in each of the 60 scenes).
Even when all five views lie within the first 40\% of the path, PSNR over the remaining 60\% (roughly 200 frames) still improves by more than $2$\,dB, as our persistent memory provides these views as conditioning for all subsequent chunks.

\subsection{Returns from a Different Viewpoint}
\label{sec:supp_returns}
\begin{table}[t]
  \centering
  \caption{\textbf{Returns from a different viewpoint.} On the directed CA-1M round trips of \cref{tab:directed}, we shift each camera on the way back by $0.5$ to $1$\,m, and report the DINOv2 similarity between its frame and the most overlapping frame from the way out (warped into its view). \name remains the most consistent method. Highlighting follows \cref{tab:directed}.}
  \label{tab:new_returns}
  \vspace{-10pt}
  \footnotesize
  \setlength{\tabcolsep}{8pt}
  \begin{tabular}{@{}l|c@{}}
    \toprule
    Method & Consistency$\uparrow$ \\
    \midrule
    AnyRecon~\cite{chen2026anyrecon} & \cellcolor{tabthird}0.566 \\
    GEN3C~\cite{ren2025gen3c} & 0.387 \\
    Lyra2~\cite{shen2026lyra} & 0.515 \\
    SolarWM-H3~\cite{huang2026SolarWM-H3} & \cellcolor{tabsecond}0.611 \\
    Spatia~\cite{zhao2026spatia} & 0.435 \\
    WorldCrafter~\cite{yu2026worldcrafter} & 0.399 \\
    \hline
    \textbf{\name} & \cellcolor{tabfirst}0.683 \\
    \bottomrule
  \end{tabular}
\end{table}
As our round trips retrace their cameras on the way back, we also evaluate consistency when each place is revisited from a different viewpoint.
To do so, we shift each camera on the way back of the directed CA-1M round trips of \cref{tab:directed} by $0.5$ to $1$\,m, while keeping it pointed at the same part of the scene.
As no two frames then share a camera, we warp the most overlapping frame from the way out into each frame of the way back (using the depth and cameras that DA3 estimates from the generated video), and measure their DINOv2 similarity over the overlap.
\name remains the most consistent method (\cref{tab:new_returns}), with a similarity of $0.683$ against $0.611$ for the next-best method (SolarWM-H3).

\subsection{Consistency vs.\ Camera Accuracy}
\label{sec:supp_metric_analyses}
\begin{table*}[t]
  \centering
  \caption{\textbf{Consistency vs.\ camera accuracy.} We recompute Revisit over only the frame pairs in which both frames have a rotation error of at most $5^\circ$, for every block of \cref{tab:directed,tab:exploration}, and report the share of its pairs that each method retains (Pairs, in \%). \name{} retains the most pairs and remains the most consistent method in every block. Highlighting follows \cref{tab:directed}.}
  \label{tab:revisit_camera}
  \vspace{-10pt}
  \setlength{\tabcolsep}{4.5pt}
  \small
  \begin{tabular}{@{}l|cc|cc|cc|cc|cc|cc@{}}
    \toprule
    & \multicolumn{4}{c|}{Directed} & \multicolumn{8}{c}{Autonomous} \\
    & \multicolumn{2}{c|}{ScanNet++} & \multicolumn{2}{c|}{CA-1M} & \multicolumn{2}{c|}{ScanNet++} & \multicolumn{2}{c|}{CA-1M} & \multicolumn{2}{c|}{DL3DV} & \multicolumn{2}{c}{SpatialVID} \\
    Method & Revisit$\uparrow$ & Pairs$\uparrow$ & Revisit$\uparrow$ & Pairs$\uparrow$ & Revisit$\uparrow$ & Pairs$\uparrow$ & Revisit$\uparrow$ & Pairs$\uparrow$ & Revisit$\uparrow$ & Pairs$\uparrow$ & Revisit$\uparrow$ & Pairs$\uparrow$ \\
    \midrule
    AnyRecon~\cite{chen2026anyrecon} & \cellcolor{tabsecond}0.730 & 26.8 & \cellcolor{tabsecond}0.701 & \cellcolor{tabthird}15.0 & \cellcolor{tabsecond}0.730 & \cellcolor{tabthird}26.8 & \cellcolor{tabsecond}0.701 & \cellcolor{tabsecond}15.0 & \cellcolor{tabthird}0.716 & 43.1 & \cellcolor{tabthird}0.724 & 30.1 \\
    GEN3C~\cite{ren2025gen3c} & 0.569 & \cellcolor{tabthird}38.8 & 0.597 & 7.9 & 0.630 & \cellcolor{tabsecond}32.9 & 0.636 & \cellcolor{tabthird}11.3 & 0.592 & \cellcolor{tabthird}58.9 & 0.581 & \cellcolor{tabsecond}36.9 \\
    Lyra2~\cite{shen2026lyra} & 0.617 & 20.4 & 0.323 & 0.3 & 0.577 & 14.4 & 0.434 & 0.2 & 0.648 & 31.8 & 0.498 & 21.8 \\
    SolarWM-H3~\cite{huang2026SolarWM-H3} & 0.470 & 4.9 & 0.539 & 3.7 & 0.474 & 7.2 & 0.464 & 7.2 & 0.478 & 25.6 & 0.468 & 16.4 \\
    Spatia~\cite{zhao2026spatia} & 0.479 & 1.7 & 0.546 & 0.6 & 0.535 & 1.5 & 0.615 & 0.7 & 0.355 & 2.0 & 0.496 & 5.3 \\
    WorldCrafter~\cite{yu2026worldcrafter} & \cellcolor{tabthird}0.693 & \cellcolor{tabsecond}45.8 & \cellcolor{tabthird}0.637 & \cellcolor{tabsecond}21.0 & \cellcolor{tabthird}0.686 & 25.9 & \cellcolor{tabthird}0.672 & 8.5 & \cellcolor{tabsecond}0.726 & \cellcolor{tabsecond}87.9 & \cellcolor{tabsecond}0.730 & \cellcolor{tabthird}36.8 \\
    \hline
    \textbf{\name} & \cellcolor{tabfirst}0.830 & \cellcolor{tabfirst}98.1 & \cellcolor{tabfirst}0.764 & \cellcolor{tabfirst}93.9 & \cellcolor{tabfirst}0.817 & \cellcolor{tabfirst}99.7 & \cellcolor{tabfirst}0.765 & \cellcolor{tabfirst}90.3 & \cellcolor{tabfirst}0.793 & \cellcolor{tabfirst}99.6 & \cellcolor{tabfirst}0.808 & \cellcolor{tabfirst}74.8 \\
    \bottomrule
  \end{tabular}
\end{table*}
As Revisit compares the two frames generated at each revisited camera, it also drops when a method deviates from the requested cameras.
To disentangle consistency from camera accuracy, we recompute Revisit over only the frame pairs in which both frames have a rotation error of at most $5^\circ$, and report the share of its pairs that each method retains (\cref{tab:revisit_camera}).
\name{} retains $75\%$ to $100\%$ of its pairs, and remains the most consistent method in every block of \cref{tab:directed,tab:exploration} ($0.76$ to $0.83$ against at most $0.73$ for the baselines), although each baseline is then scored only on the pairs in which it follows the cameras.
Its lead over every baseline is reliable in every block (the paired $95\%$ confidence intervals over scenes exclude zero), and persists with a threshold of $10^\circ$.

\subsection{Layout Ablation}
\label{sec:supp_layout_variants}
\begin{table*}[t]
  \centering
  \caption{\textbf{Layout ablation.} We replace the planned layout on the directed round trips of \cref{tab:directed} with only the four walls of a room that encloses the observed points and the cameras, or with the described objects at random positions within this room. We also report the fraction of each box that its named object fills (Label adh., \cref{sec:supp_metrics}) and the fraction of the described spatial relations that the layout satisfies (Relations). Coarse geometry already anchors the camera, but only the planner's layout determines what appears where, and it performs best on every metric.}
  \label{tab:proxies}
  \vspace{-10pt}
  \small
  \setlength{\tabcolsep}{0pt}
  \begin{tabular*}{\linewidth}{@{\extracolsep{\fill}}lcccccccccc@{}}
    \toprule
    & \multicolumn{5}{c}{ScanNet++~\cite{yeshwanth2023scannetplusplus}} & \multicolumn{5}{c}{CA-1M~\cite{lazarow2025cubify}} \\
    \cmidrule(lr){2-6}\cmidrule(l){7-11}
    Layout & RotErr$\downarrow$ & TransErr$\downarrow$ & Revisit$\uparrow$ & Label adh.$\uparrow$ & Relations$\uparrow$ & RotErr$\downarrow$ & TransErr$\downarrow$ & Revisit$\uparrow$ & Label adh.$\uparrow$ & Relations$\uparrow$ \\
    \midrule
    w/o layout & 5.3 & 0.08 & 0.785 & 0.144 & -- & 14.6 & 0.22 & 0.653 & 0.166 & -- \\
    Walls only & 3.3 & 0.12 & 0.788 & -- & 0.000 & 3.4 & 0.16 & 0.720 & -- & 0.000 \\
    Randomly placed objects & 1.8 & 0.06 & 0.807 & 0.122 & 0.252 & 3.7 & 0.12 & 0.731 & 0.083 & 0.305 \\
    \textbf{\name} & \textbf{1.3} & \textbf{0.04} & \textbf{0.826} & \textbf{0.298} & \textbf{0.608} & \textbf{2.2} & \textbf{0.07} & \textbf{0.755} & \textbf{0.278} & \textbf{0.559} \\
    \bottomrule
  \end{tabular*}
\end{table*}
Our layout serves two purposes: it anchors the camera to its path in unseen regions, and it specifies what appears where.
To measure what each purpose requires, we replace the planned layout on the directed round trips of \cref{tab:directed} with two layouts that require no planner: the four walls of a room that encloses the observed points and the cameras, or the described objects at typical sizes and random positions within this room (\cref{tab:proxies}).

Although neither involves the planner, both already anchor the camera (\eg, reducing the rotation error on CA-1M from $14.6^\circ$ without a layout to $3.4^\circ$ and $3.7^\circ$), as even coarse geometry conveys the camera motion once rendered along its path.
This is in line with our key insight that the interface between the two models can be coarse, and with the robustness of \name to imprecise boxes (\cref{sec:supp_planning}) and smaller planners (\cref{sec:supp_planners}).
However, neither alternative places the described objects where the descriptions put them.
Walls contain no object, and the named object fills less than half as much of a randomly placed box as of a planned one (label adherence of $0.12$ and $0.08$ against $0.30$ and $0.28$), while such a layout satisfies only $25\%$ and $31\%$ of the spatial relations that the descriptions state (against $61\%$ and $56\%$).
The planner's layout performs best on every metric on both datasets, and follows the cameras more accurately than the walls ($1.3^\circ$ against $3.3^\circ$ on ScanNet++, and $2.2^\circ$ against $3.4^\circ$ on CA-1M).
Thus, coarse geometry anchors the camera, but only the planner's layout determines what appears where.

\subsection{Qualitative Ablation}
\label{sec:supp_ablation_example}
\begin{figure*}[t]
    \centering
    \includegraphics[width=\linewidth]{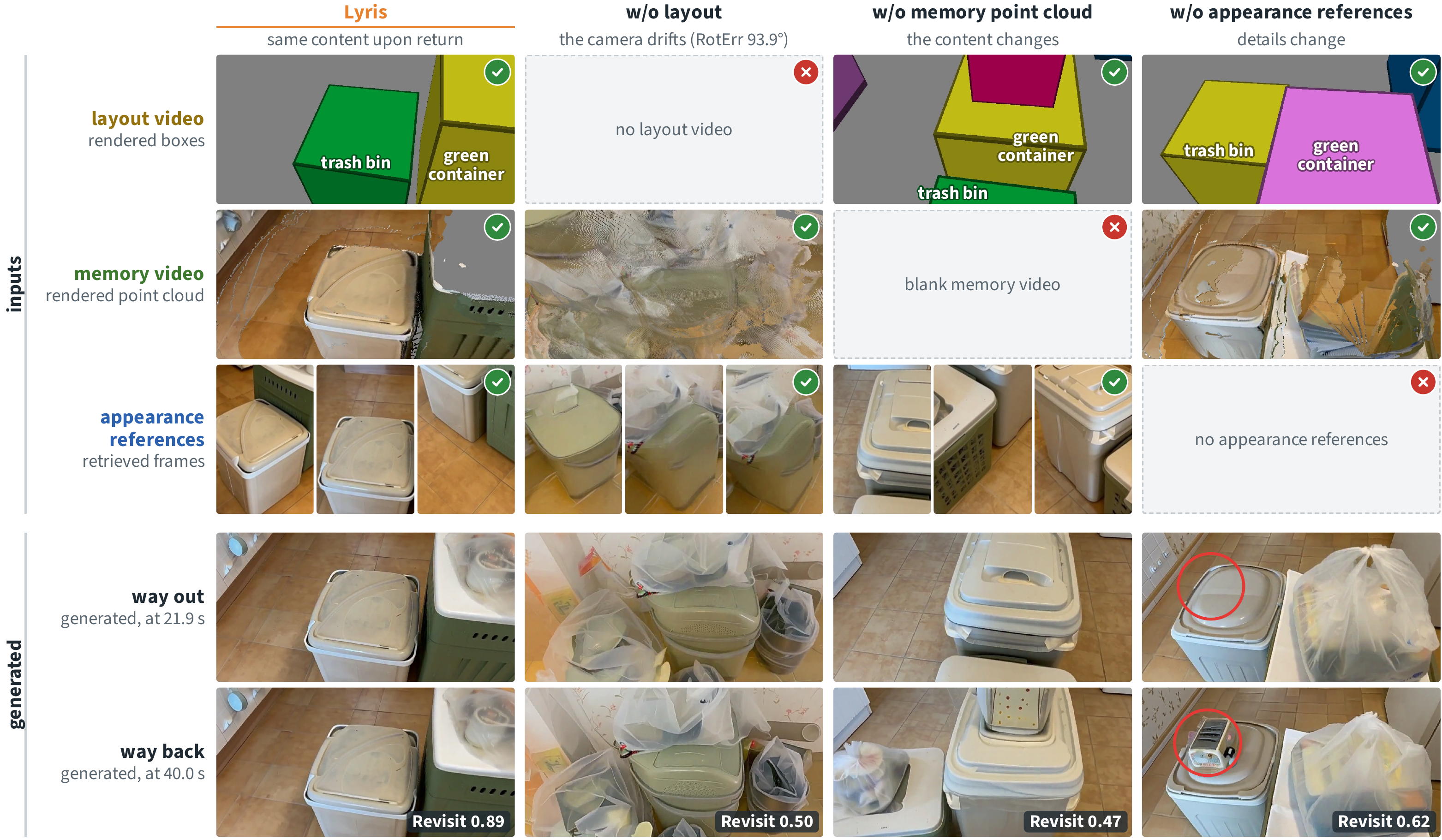}  %
    \vspace{-15pt}
    \caption{\textbf{Qualitative ablation.} \name and its ablations (\cref{tab:diagnostics}) on a directed CA-1M round trip. Top: each variant's inputs (check: given, cross: removed). Bottom: the frames generated at the same camera at 21.9\,s and 40.0\,s, with their DINOv2 similarity (Revisit). \name returns to the same content, whereas the camera drifts without the layout (RotErr of $93.9^\circ$ at this camera), the content changes without the memory's point cloud, and details change without the appearance references (circled).}
    \label{fig:diagnostics}
    \vspace{-10pt}
\end{figure*}
We illustrate the ablations of \cref{tab:diagnostics} in \cref{fig:diagnostics}, at a camera that a directed CA-1M round trip visits on the way out and again on the way back.
\name generates the same content upon return.
Without the layout, the camera drifts from its path (with a rotation error of $93.9^\circ$ at this camera, against $2.0^\circ$ for \name) and views a different part of the scene.
Without the memory's point cloud (and thus with a blank memory video), the appearance references alone do not preserve the scene, and the content changes upon return.
Conversely, the point cloud alone preserves the overall structure, but details change without the appearance references (\eg, the circled object that appears on the lid).

\subsection{Layout Editing}
\label{sec:supp_motion}
As our layout consists of one labeled box per object, a user can directly edit it to control what \name generates.
In unseen regions, the boxes determine what is generated and where, and moving, swapping, or adding them changes the content accordingly (\cref{fig:why_layouts}).
In the existing scene, objects are also stored in our memory, which we thus edit along with the layout.
As this memory is an explicit point cloud, we simply move an object's points together with its box, and hide the points that a new box occludes.
Although \name is trained only on static scenes, we find that moving a box over time also moves its object in the generated video (\eg, a chair sliding away from a wall or a ball rolling across the floor).
We illustrate each of these edits on our project page.

\section{Method Details}
\label{sec:supp_method}
\input{sec_clean/X_suppl_method}

\section{Evaluation Protocol}
\label{sec:supp_eval}

\subsection{Evaluation Scenes}
\label{sec:supp_scenes}
We select the evaluation scenes of \cref{tab:directed,tab:exploration} from the captured footage and its metadata alone, independently of any method's output.
For CA-1M, we use the 20 longest landscape captures, and for ScanNet++, we randomly sample 20 of the held-out captures after excluding those with very fast camera motion, as our task is a walkthrough at a natural pace from a single image.
For DL3DV, we use DL3DV-Evaluation~\cite{ling2024dl3dv}, a separately released test split\footnote{\texttt{DL3DV/DL3DV-Evaluation} on Hugging Face.} that does not overlap with the DL3DV-10K scenes we train on.
From this split, we likewise exclude the scenes with very fast camera motion, as well as those whose start frame is dominated by dense vegetation, which leaves no clear path to walk through.
We then randomly sample 20 of the remaining scenes (16 indoor and 4 outdoor).
As the annotated clips of SpatialVID~\cite{wang2025spatialvid} last at most 15\,s, which is too short for our round trips, we instead use its long raw videos (one clip per video) and estimate their cameras with ViPE~\cite{huang2025vipe}.
From these in-the-wild videos, we only keep ground-level walkthroughs of real, static scenes without overlaid text.
To ensure that camera errors are meaningful, we further require that DA3~\cite{lin2025depth}, with which we measure them, recovers the ViPE cameras from the captured footage to within $10^\circ$.
For a fair comparison, we also require enough sparse points in the start frame to calibrate the scale of the input cameras (\cref{sec:supp_baselines}), and exclude the clips that SolarWM-H3~\cite{huang2026SolarWM-H3} is trained on.
The 20 videos that pass all criteria (17 indoor and 3 outdoor) form our test set.

\subsection{Metrics}
\label{sec:supp_metrics}
We score each method from its generated frames alone, without access to its internal cameras or geometry.
As the way back retraces the way out, we measure Revisit across the two and all other metrics on the way out, and average each over scenes.

\topic{Camera accuracy.}
As every method starts from the same frame, we align the cameras that DA3\footnote{\texttt{DA3NESTED-GIANT-LARGE-1.1} on Hugging Face.} recovers from each generated video with the input cameras at this frame only, such that the errors include any drift along the path, and fit a single scale to the recovered positions.
RotErr is then the mean per-frame angle between the rotations of the recovered and the input cameras (in degrees), and TransErr is the mean distance between their positions (in the units of the input cameras, which are metric on ScanNet++ and CA-1M).

\topic{Observed pixels.}
To restrict PSNR, SSIM, and LPIPS to the pixels that the start frame observes, we determine these pixels from the capture, such that they are identical for every method.
A pixel is observed if its captured surface reprojects into the start frame within a relative depth tolerance of 10\%, using the captured depth of CA-1M and ScanNet++ and MoGe-3~\cite{kong2026moge} depth scaled to the sparse points of DL3DV and SpatialVID.

\topic{Consistency upon return.}
Revisit compares the two frames generated at each revisited camera directly, as the cosine similarity of their DINOv2-large patch features.
We exclude the cameras that are revisited within 48 frames (around the turn), as frames this close in time remain consistent without memory.

\topic{Objects.}
As the judge should favor no method, we use Gemini 3.1 Pro at temperature 0 (a model that neither plans our layouts nor writes the descriptions), and withhold the name of the method.
We ask it which of the objects that each description names (as extracted by GPT-6) appear in 31 evenly spaced frames of the way out.
This judge finds 97.6\% and 97.8\% of the objects in the captured videos of ScanNet++ and CA-1M, and returns the same answer for 95.8\% of the objects when queried a second time.

\topic{Layout adherence.}
For \cref{tab:layout}, we obtain the generated surface from the DA3 depth of the generated frames.
Adherence measures whether objects are generated where the layout places them, as the fraction of the pixel rays through a box at which this surface lies inside the box instead of behind it.
As a box only bounds its object, the captured objects themselves fill about 0.6 of their annotated boxes on both datasets, so the adherence of \name to the annotated layout in \cref{tab:layout} is at the level of the real scene.
Recall measures whether the real objects are reproduced, as the fraction of the pixels showing an annotated object in the capture at which the generated surface also lies inside its box.
We average both over the boxes that the start frame does not observe, as the input shows the others.

\topic{Label adherence.}
As adherence does not test what fills a box, we further report the label adherence in \cref{tab:proxies}, as the fraction of the visible pixels of each box that SAM~3~\cite{carion2026sam3} segments when prompted with the box's label.
We also average over the boxes that the start frame does not observe, and score the videos generated without a layout on the boxes of the planned layout.
On the captured videos with their annotated boxes, it reaches $0.39$ and $0.35$ on ScanNet++ and CA-1M (as a box also covers what lies behind its object), against $0.03$ and $0.04$ when we shuffle the labels within each scene.

\subsection{Baselines}
\label{sec:supp_baselines}
We run the official implementation of each baseline with its default settings (\cref{tab:baseline_settings}), and generate each round trip in a single run with a fixed seed, so that every method relies on its own memory.
As the baselines generate between 17 and 121 frames at a time, we provide each step with the description of the 124-frame window of the path that contains its middle frame (or the caption of the start frame in the autonomous setting).
AnyRecon is the only exception, as its official implementation uses an empty prompt, so its directed and autonomous round trips coincide.
To ensure that all methods receive the same camera path, we calibrate the scale of the input cameras once per scene from the sparse points visible in the start frame, and use it for every method.
Each baseline generates at its default resolution (at least $832\times480$), and we downsample all videos to $832\times480$ for evaluation, which, if anything, favors the baselines that generate at a higher resolution.
\begin{table}[t]
  \centering
  \caption{\textbf{Generation settings.} We list the number of frames that each method generates per step, its resolution, and the seed of its single run per round trip. All baselines use the default settings of their official implementations, and generate at the resolution of \name or higher.}
  \label{tab:baseline_settings}
  \vspace{-10pt}
  \footnotesize
  \setlength{\tabcolsep}{0pt}
  \begin{tabular*}{\linewidth}{@{\extracolsep{\fill}}lccc@{}}
    \toprule
    Method & Frames per step & Resolution & Seed \\
    \midrule
    AnyRecon~\cite{chen2026anyrecon} & 35 & $896\times512$ & 1 \\
    GEN3C~\cite{ren2025gen3c} & 121 & $1280\times704$ & 1 \\
    Lyra2~\cite{shen2026lyra} & 80 & $832\times480$ & 1 \\
    SolarWM-H3~\cite{huang2026SolarWM-H3} & 17 & $1344\times768$ & 42 \\
    Spatia~\cite{zhao2026spatia} & 121, then 81 & $1248\times704$ & 1 \\
    WorldCrafter~\cite{yu2026worldcrafter} & 33 & $832\times480$ & 1 \\
    \hline
    \textbf{\name} & 124 & $832\times480$ & 42 \\
    \bottomrule
  \end{tabular*}
\end{table}

\section{Limitations}
\label{sec:supp_failures}
As \name builds its memory from estimated geometry, depth and alignment errors can accumulate over very long paths.
Our acceptance test (\cref{sec:supp_memory_update}) discards unreliable updates, and stronger reconstruction models are directly applicable to our method.
\name does not currently model dynamic scenes, as it is trained only on static captures.
However, moving a box over time already moves its object in the generated video (\cref{sec:supp_motion}), which makes training on dynamic data a promising direction.
Our layout specifies only a labeled box per object, and thus offers no control over its detailed shape, which richer primitives rendered into the layout video could provide.
Finally, \name is not yet interactive, as planning and sampling run in sequence for every chunk, which faster planners and distillation into a causal model~\cite{de2026artifixer} could address.

\section{Societal Impact}
\label{sec:supp_impact}
\name generates explorable environments from as little as a single photo, which benefits applications such as virtual touring, content creation, and simulation for physical AI.
Its explicit layout further allows a user to inspect and edit what will be generated before any frame is rendered.
However, as \name plausibly extends real places beyond what was captured, it raises concerns about potential misuse for generating deceptive visual media (\eg, presenting a generated walkthrough as real footage).
Although \name is trained on static scenes and is not designed to synthesize people, appropriate safeguards such as watermarking generated content should be considered for deployment.
As \name builds on a pretrained video model and a vision-language planner, it also inherits the biases of their training data (\eg, toward common room types and furnishings).

\ifdefined\putbib{\small\putbib[main]}\end{bibunit}\fi

%% file: sec_clean/X_suppl_method.tex
\subsection{Layout Planning}
\label{sec:supp_planning}
\begin{figure}[htbp]
\footnotesize
\begin{verbatim}
{"objects": [
  {"label": "aquarium",
   "center": [2.03, 1.47, 0.6],
   "size": [1.0, 0.4, 1.2], "yaw": 155.7},
  {"label": "hookah",
   "center": [1.35, 1.78, 0.4],
   "size": [0.3, 0.3, 0.8], "yaw": 155.7},
  {"label": "orchid",
   "center": [1.76, 1.59, 1.425],
   "size": [0.35, 0.25, 0.45], "yaw": 155.7}]}
\end{verbatim}
\vspace{-10pt}
\caption{\textbf{Planner output.} We show three of the boxes that the planner adds in the second chunk of a CA-1M round trip, without their colors and justifications. Each box is fully specified by a label, a center and size in meters, and a rotation about the gravity axis in degrees.}
\label{fig:planner_output}
\vspace{-10pt}
\end{figure}
\topic{Planning.}
We use Claude Opus 5.5 as our planner, although \name also outperforms all baselines with smaller and open-weight planners (\cref{sec:supp_planners}).
As our layout only requires one labeled box per object, we simply query the planner with fixed prompts, without tools or fine-tuning.
It first localizes each object visible in the start frame with a 2D bounding box, which we lift to 3D by fitting a box to the memory points within it.
To place the objects that the chunk should reveal, we render the point-cloud memory and the fitted boxes from above and from four of the chunk's cameras, overlaid with a metric grid so that the planner can specify each new box in metric coordinates (\cref{fig:planner_output}).
Similar to LayoutGPT~\cite{feng2023layoutgpt}, we retrieve two layouts with matching objects from the annotated ScanNet++ scenes of our training set as in-context examples, and provide median object sizes.

\begin{table}[t]
\centering
\footnotesize
\caption{\textbf{Geometric checks.} We verify each box that the planner proposes with the six checks of \cref{sec:layout}. Their tolerances are the same for every scene and dataset.}
\vspace{-8pt}
\begin{tabular}{@{}lp{0.76\linewidth}@{}}
\toprule
Check & A box passes if \\
\midrule
Support & it rests on the floor (within $0.15$\,m), on another box, or on the memory's points, or hangs on a wall \\
Floor plan & its center lies inside the floor plan \\
Free space & at most 25\% of it lies in space that the start frame observes to be empty \\
Collisions & at most 5\% of the smaller of two boxes lies inside the other, or no more than for annotated objects of their classes in our training scenes (\eg, a chair under a table) \\
Visibility & at least $2{,}000$ of its pixels ($500$ for objects smaller than $0.6$\,m) are unoccluded in at least 20\% of the chunk's frames \\
Relations & it satisfies the spatial relations that its description states \\
\bottomrule
\end{tabular}
\label{tab:checks}
\vspace{-10pt}
\end{table}
\topic{Verification.}
We sample five proposals and verify each with the six geometric checks of \cref{sec:layout}, whose criteria we list in \cref{tab:checks}.
We select the proposal with the fewest violations and let the planner revise it once, after which the deterministic repair resolves the remaining violations by displacing each flagged box as little as possible.
Finally, we remove the placed boxes that are still not visible, collide with another box, or contradict a stated relation.
The checks use the same tolerances for every scene and dataset. They are mainly for gross errors, and their exact values are not critical, as \name is robust to imprecise boxes.
Even when we perturb the final boxes with Gaussian noise (standard deviations of up to $0.3$\,m, 25\%, and $20^\circ$ in position, size, and orientation), the rotation error on the directed CA-1M round trips increases by less than $1^\circ$, and \name still outperforms all baselines on every metric of \cref{tab:directed}.

\topic{Autonomous setting.}
Given only a caption of the input view, the planner itself decides which objects each chunk should reveal, places and verifies them as above, and writes the chunk's description from the resulting layout (the Layout as text variant of \cref{tab:diagnostics} receives this description without the layout video).
When the full camera path is available before generation (\name w/ full path in \cref{tab:exploration}), the planner instead plans the layout for the whole path at once, from the input views and the cameras of all chunks.
Planning chunk by chunk does not require the whole path in advance,
which enables open-ended exploration, and \name still
outperforms all baselines in this setting
(\cref{tab:exploration}).
On CA-1M and SpatialVID, where three of the 40 round trips account for most of the difference in rotation error between the two variants (\cref{tab:exploration}), the median over scenes is $1.9^\circ$ and $2.2^\circ$ when planning chunk by chunk, against $1.8^\circ$ on both with the full path.

\subsection{Layout Rendering and Update}
\label{sec:supp_layout_update}
\topic{Rendering.}
As the text prompt refers to each box by its color, we render the boxes of each training clip in perceptually distinct named colors, which we sample at random for each clip.
As each training clip contains at most 12 boxes, we likewise render at most 12 boxes per chunk at inference, and sample a palette of 12 colors once per path.
Each label thus retains its color across chunks, and a path with more than 12 labels reuses colors across chunks while keeping those within each chunk distinct.

\topic{Update.}
As the memory video already shows what has been generated, the layout video prioritizes rendering only what has not.
Before each subsequent chunk, we thus measure how much of each object the memory contains, as the fraction of its box's pixels that are covered by memory points inside the box.
Above 0.8, we consider the object fully generated and exclude its box.
Between 0.1 and 0.8, we consider it partially generated and refit its box to the largest cluster of memory points around it, so that the remainder of the object continues the part that exists.
As this cluster can include neighboring surfaces (\eg, the wall behind the object), we preserve the orientation of the box and allow each side to grow by at most $1.5\times$.
Otherwise, the object is yet to be generated and its box remains as planned.
When more than 12 boxes remain, we prioritize those planned in unseen regions over those fitted to visible objects, as the prefix frame already shows the latter.

\subsection{Reference Retrieval}
\label{sec:supp_retrieval}
From the bank $\mathcal{H}$ of input views and previously generated frames, we retrieve up to three appearance references per chunk, selected to maximize their joint coverage of the chunk's views.
For each frame $i$ in $\mathcal{H}$, we project its estimated depth into every eighth camera of the chunk to obtain the set $\mathcal{V}_i$ of pixels that it observes, with a depth test for occlusion.
As each chunk already receives its prefix frame, we initialize the selection $\mathcal{R}$ with it and greedily add the frame that maximizes the additional coverage,
\begin{equation}
i^{*}
=
\underset{i\notin\mathcal{R}}{\arg\max}
\bigl|
\mathcal{V}_i
\setminus
\textstyle\bigcup_{j\in\mathcal{R}}\mathcal{V}_j
\bigr|.
\label{eq:reference_retrieval}
\end{equation}
Once no frame increases the coverage, we select the remaining references by the proximity of their cameras to the chunk's.
When a chunk returns close to the start of the path, we additionally include the first input view, so that the chunk is conditioned on the captured view itself.

\subsection{Memory Update}
\label{sec:supp_memory_update}
We initialize the memory with the monocular depth~\cite{kong2026moge} of the first input view, and reconstruct any further input views jointly with Pi3X~\cite{wang2026pi, pi3x}.
As errors in this memory would persist for the rest of the video, we align each chunk to it, discard the update altogether if the aligned cameras deviate from the requested ones, and otherwise add only the points that are reliable and not yet contained in it.

\topic{Alignment.}
We reconstruct every eighth frame of the chunk with Pi3X, conditioned on the requested cameras, together with frames that the memory already contains, and fit a similarity transform~\cite{umeyama1991least} to the 3D points of these shared frames.
With a single input view, the first chunk shares no reconstructed frame with the memory, and we instead fit the transform to its camera centers.
As these alone can misestimate the scale, we then rescale the reconstruction to the depth of the input view.

\topic{Acceptance test.}
We accept the update only if the median rotation between the aligned and the requested cameras is at most $15^\circ$ and the root-mean-square error of their centers is at most $0.25\max(s, 0.2\,d)$, where $s$ is the cameras' spread (the root-mean-square distance of their centers from their mean) and $d$ is the median depth of the new frames.
The term $0.2\,d$ is needed because $s$ vanishes when the camera turns in place.
On the directed CA-1M round trips, 93\% of the updates pass this test, against 78\% without the layout, as the generated cameras then follow the requested ones less closely (\cref{tab:diagnostics}).

\topic{Filtering.}
From each reconstructed frame of an accepted update, we first discard the points with low Pi3X confidence or at depth discontinuities, where the reconstruction is least reliable.
To keep a single copy of each surface, we then render the current point cloud into the frame's camera and discard the points whose depth agrees with the rendered depth within a relative tolerance.
We likewise discard the points that lie on or in front of the surfaces observed by the first input view, as they would duplicate or contradict this captured view, and add the remaining points to $\mathcal{P}$.
To ensure that the memory video still covers the entire prefix frame of the next chunk, we also lift this frame with monocular depth, scaled to the memory, and render its points wherever the memory provides none.

\subsection{Chunk Transitions}
\label{sec:supp_transitions}
\topic{Chunks.}
We generate 124 frames per chunk and keep the first 107 (all but the last latent group of the video VAE, which has the lowest imaging quality). Each chunk continues from the last frame that we keep of the previous one, and the last chunk ends at the last frame of the path and is kept in full, such that a 744-frame round trip consists of seven chunks.
We plan a layout (\cref{sec:supp_planning}) before each chunk whose cameras no earlier chunk has covered. On a round trip, these are the first four of the seven chunks, and the remaining three retrace the way out, so they add no boxes and reuse the description of the chunk that they retrace.
In the directed setting, each planned chunk receives the description of the 124-frame window of the path that contains its middle frame, as does each step of the baselines (\cref{sec:supp_baselines}).

\topic{Transitions.}
Each chunk continues from the last frame that we keep of the previous one.
As the model is trained with captured prefix frames, we sharpen this generated frame with an unsharp mask (radius of 2 pixels, amount of 70\%) before it conditions the next chunk, while the output video retains the original frame.
In the first chunk, we also replace the latents that the memory video covers with its own (noised to the current level) after each of the first two of the eight sampling steps, which keeps the chunk aligned with the input views.
These choices slightly improve consistency upon return and imaging quality (by 0.026 and 1.5 on the directed CA-1M round trips of \cref{tab:directed}, and by 0.021 and 1.4 on those of ScanNet++).
Without them, \name still outperforms all baselines in camera accuracy, consistency upon return, and imaging quality.

\subsection{Architecture}
\label{sec:supp_architecture}
\begin{figure}[t]
  \centering
  \includegraphics[width=0.72\linewidth]{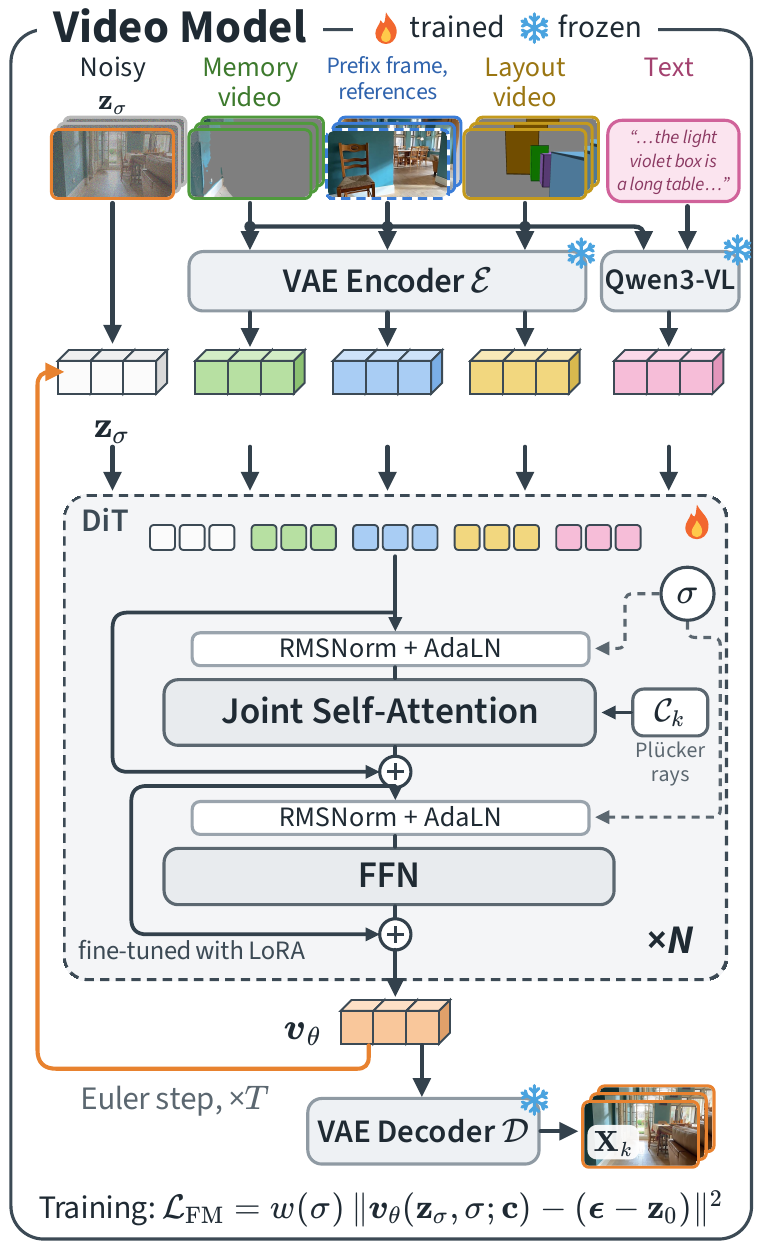}
  \caption{\textbf{Video model.} The DiT processes the latents of the noisy target together with those of the memory video, the prefix frame and appearance references, and the layout video, and with the Qwen3-VL embedding of the text prompt, through joint self-attention. We freeze the VAE and Qwen3-VL, fine-tune the DiT with LoRA, and add only a zero-initialized linear layer per block that embeds the cameras $\mathcal{C}_k$ as Pl\"ucker rays.}
  \label{fig:architecture}
  \vspace{-20pt}
\end{figure}
To preserve the prior of the pretrained model, we keep the architecture of MiniMax-H3 Ref2VA model unchanged apart from adding the camera embedding, and provide all other conditions through its native reference inputs (\cref{fig:architecture}).
The VAE latents of these conditions thus enter the joint self-attention of the DiT as separate tokens, while the Qwen3-VL encoder embeds their frames together with the text prompt.
As the prefix frame is the first frame of the chunk, its tokens share the RoPE positions of the first target frame, whereas the appearance references keep separate positions.
For the camera embedding, we compute one Pl\"ucker ray per latent position, relative to the first target camera, averaging the cameras over the frames that each latent frame compresses.
Each target and reference-image token receives its rays at the self-attention input of every transformer block, through a linear layer that we zero-initialize so that training starts from the pretrained model.

\subsection{Training and Distillation}
\label{sec:supp_training}
\topic{Training.}
The two stages of \cref{sec:setup} expose the model to each situation that it faces at inference.
Both stages use the optimizer of \cref{sec:setup}, with gradient clipping at $1.0$ and one clip per GPU.
The first stage ($1{,}500$ iterations on 128 GPUs) covers first chunks, with an even mix of ScanNet++ clips and DL3DV clips that start from one to five input views.
In the second stage ($500$ iterations on 64 GPUs), 42.5\% of the clips continue from a captured previous chunk, from which we build their memory and retrieve their references as at inference.
To cover returns, 20\% play this previous chunk back in reverse and 12.5\% do the same for an earlier chunk.
The remaining 25\% are first-stage clips, which keeps first chunks in the training distribution.

\topic{Condition dropout.}
We represent a memory video that is dropped during training (\cref{sec:training}) by the VAE encoding of an all-gray sequence, as for a memory without points.
We omit a dropped layout video together with the part of the prompt that names its boxes.
When we drop the text prompt, we retain the Qwen3-VL embeddings of the reference images and videos, so that classifier-free guidance acts on the text prompt alone.

\topic{Distillation.}
To sample with only eight steps and without classifier-free guidance, we distill the second-stage model with DMD2~\cite{yin2024improved} using the same training clips for $500$ iterations on 32 GPUs, using it as the teacher at a guidance scale of 3.
Every fifth iteration updates the student (learning rate $10^{-5}$), and the others update the fake score model (a LoRA, learning rate $10^{-4}$).